# Meta-Inverse Physics-Informed Neural Networks for High-Dimensional Ordinary Differential Equations

Zhao Wei, Kenneth Hor Cheng Koh, Sheng Yuan Chin, James Chun Yip Chan, Chin Chun Ooi, and Yew-Soon Ong, *Fellow, IEEE*

***Abstract*—Solving inverse problems in dynamical systems governed by high-dimensional coupled ordinary differential equations (ODEs) is a ubiquitous challenge in scientific machine learning. In many real-world applications, researchers seek to uncover unknown parameters or model unknown dynamics even as the underlying physics is only partially characterized, and observations are sparse and limited to specific measurable channels. While physics-informed neural networks (PINNs) are ideal for inverse inference under partial observability, existing PINNs typically rely on task-specific joint optimization, which suffers from optimization difficulties and poor generalization. In this paper, we propose a meta-inverse physics-informed neural network (MI-PINN) that reformulates inverse modeling as a two-stage meta-learning problem. MI-PINN first learns a physics-aware representation across multiple tasks, and then performs inverse modeling by optimizing task-specific unknowns while keeping the learned representation fixed. This two-stage formulation significantly reduces the parameter search dimension, thereby improving sample efficiency and enabling accurate inference. To handle multi-scale dynamics common in these high-dimensional ODE systems, we further introduce an adaptive clustering-based multi-branch learning scheme. We demonstrate the effectiveness of MI-PINN on whole-body physiologically based pharmacokinetic (PBPK) models with up to 33 coupled ODEs, using paracetamol and theophylline under intravenous and oral dosing scenarios. Experimental results show that MI-PINN enables accurate recovery of masked kinetic parameters and reconstruction of missing mechanistic terms despite limited clinical observations. Compared with conventional solver-based optimization and standard inverse PINN baselines, MI-PINN consistently achieves up to two-orders-of-magnitude reductions in parameter estimation error while improving computational efficiency. These results highlight MI-PINN as a generalizable framework for data-efficient inverse inference and discovery of unknown mechanisms in real-world high-dimensional coupled ODE systems, including for whole-body PBPK with potential application to drug discovery.**

## I. Introduction

INVERSE learning in high-dimensional dynamical systems is a fundamental challenge in scientific machine learning [1]. In many real-world applications, unknown physical parameters or missing mechanisms must be inferred from sparse and noisy observations, even as system dynamics are governed by high-dimensional, tightly coupled nonlinear ordinary differential equations (ODEs) [2]. This problem is compounded by the fact that many such systems exhibit only partial observability, i.e., measurements occur in a limited subset of the physical state variables. These inverse problems are often ill-posed, highly non-convex, and strongly sensitive to model structure and data availability, making stable and reliable inference particularly difficult [3]. Hence, pure data-driven learning methodologies struggle when applied to such complex systems, frequently overfitting to the sparse data and/or generating physically inconsistent predictions [4].

Physics-informed neural networks (PINNs) have recently emerged as a promising alternative for solving inverse problems governed by differential equations [5-8]. By embedding known physical laws directly into the training objective, PINNs enable the simultaneous learning of system states, unknown parameters, and even missing dynamical components, while respecting the underlying mechanistic constraints [9]. However, existing PINN-based approaches largely formulate inverse modeling as a task-specific joint optimization problem, where neural network parameters and unknown physical variables are optimized simultaneously for each individual system [5, 10]. This formulation becomes increasingly problematic in high-dimensional ODE systems,

This research was partially supported by the A*STAR Catalyst Project for Artificial Intelligence in Drug Discovery (AIDD) Programme (Grant No. H25MHP1489) and the A*STAR under BMRC CSF (Grant No. C240314053).

Zhao Wei is with the Centre for Frontier AI Research (CFAR) and the Institute of High Performance Computing (IHPC), Agency for Science, Technology and Research (A*STAR), 1 Fusionopolis Way, #16-16 Connexis, Singapore 138632, Singapore (e-mail: Wei_Zhao@a-star.edu.sg).

Kenneth Hor Cheng Koh and Sheng Yuan Chin are with the Singapore Institute of Food and Biotechnology Innovation (SIFBI), Agency for Science, Technology and Research (A*STAR), 31 Biopolis Way, Nanos, Singapore 138669, Singapore (e-mail: Koh_Hor_Cheng@a-star.edu.sg, Chin_Sheng_Yuan@a-star.edu.sg).

James Chun Yip Chan is with the Singapore Institute of Food and Biotechnology Innovation (SIFBI), Agency for Science, Technology and Research (A*STAR), 31 Biopolis Way, Nanos, Singapore 138669, Singapore and the A*STAR Skin Research Labs (A*SRL), Agency for Science, Technology and Research (A*STAR), 8A Biomedical Grove, Immunos, Singapore 138648, Singapore (e-mail: James_Chan@a-star.edu.sg).

Chin Chun Ooi (Corresponding author) is with the Centre for Frontier AI Research (CFAR) and the Institute of High Performance Computing (IHPC), Agency for Science, Technology and Research (A*STAR), 1 Fusionopolis Way, #16-16 Connexis, Singapore 138632, Singapore (e-mail: ooicc@a-star.edu.sg)

Yew-Soon Ong is with the Centre for Frontier AI Research (CFAR) and the Institute of High Performance Computing (IHPC), Agency for Science, Technology and Research (A*STAR), 1 Fusionopolis Way, #16-16 Connexis, Singapore 138632, Singapore and the College of Computing and Data Science (CCDS), Nanyang Technological University (NTU), 50 Nanyang Avenue, Singapore 639798, Singapore (e-mail: asysong@ntu.edu.sg).

where strong coupling and multi-scale dynamics and state behaviors lead to highly complex loss landscapes and poor optimization stability [11]. In addition, these approaches exhibit limited generalization across tasks. Models trained on one system or parameter setting typically need to be retrained from scratch and cannot be directly reused for new tasks, even when they share similar underlying physics [12]. Consequently, their scalability for inverse design and discovery in complex dynamical systems remains limited.

These limitations suggest that inverse learning should move beyond task-specific optimization and instead exploit shared structures across related dynamical systems. In many practical settings, different systems can be viewed as variations of a common underlying process, sharing similar dynamics but differing in partially characterized parameters or mechanisms. Meta-learning provides a natural framework to capture such shared structure by learning representations that can be reused across tasks, improving data efficiency and generalization [13]. However, its application to physics-informed inverse problems remains limited, especially in high-dimensional ODE systems. The tight coupling between representation learning and physics-constrained optimization, together with multi-scale dynamics, makes it difficult to obtain generalizable solutions.

To address these challenges, we propose a meta-inverse physics-informed neural network (MI-PINN) that reformulates inverse learning in high-dimensional coupled ODE systems as a two-stage learning problem. By decoupling representation learning and inverse modeling, MI-PINN reduces the optimization space and supports efficient cross-task adaptation. To rigorously evaluate the proposed MI-PINN framework on a real-world complex and high-dimensional ODE system, we consider whole-body physiologically based pharmacokinetic (PBPK) models. While traditionally studied in pharmacology, mathematically, whole-body PBPK systems represent an exceptionally demanding benchmark for scientific machine learning [14]. They consist of multiple tightly coupled ODEs characterized by varying temporal scales, highly nonlinear interaction terms, and extreme partial observability. In practice, the ODE system can contain more than 100 equations, even as only a single state variable (such as venous blood concentration) is directly measurable. Successfully performing parameter inference and/or neural-symbolic missing term discovery under such conditions is not only challenging, but also carries real-world practical implications. Drug development remains a costly and protracted endeavour, with poor pharmacokinetics (PK) being a major contributor to the high attrition rate in drug development, accounting for approximately 40% of clinical failures [15, 16], making accurate inverse inference from limited clinical observations especially compelling [17].

The main contributions are listed as below.

1) We propose a MI-PINN framework that decouples representation learning and inverse inference as a two-stage formulation, enabling data-efficient inverse inference under limited data and partial observability. This transforms inverse learning in high-dimensional ODE systems from a joint optimization problem with search space dimension $N_\theta$ (network weights) + $N_\alpha$ (task-specific unknowns) into a substantially reduced inference problem with dimension $N_\alpha$ ($N_\theta >> N_\alpha$). In addition, MI-PINN achieves accurate inverse modeling with as few as 10 observations from a single state variable, demonstrating improved sample efficiency in high-dimensional settings.

2) A multi-branch representation scheme is proposed to capture multi-scale dynamics common in coupled ODE systems through an adaptive clustering-based learning strategy. This is crucial to overcome optimization difficulties due to spectral bias that complicate PINN optimization in the presence of multi-scale dynamics. In addition, we introduce a physics-informed pseudo-inverse formulation that computes the final layer in closed form to improve training stability.

3) The proposed MI-PINN framework provides a unified approach for both parameter inference and missing mechanism discovery from limited observations in inverse modeling. We further demonstrate the practical scalability of the MI-PINN framework on complex, high-dimensional whole-body PBPK models with up to 33 ODEs under intravenous (IV) and oral dosing scenarios. Experiments on paracetamol and theophylline consistently yield accurate parameter inference under sparse and noisy observations, as well as successful reconstruction of missing mechanistic terms at hitherto unprecedented scale.

The rest of this paper is organized as follows. Section II reviews related work. Section III presents the proposed MI-PINN framework. Section IV introduces its application to inverse PBPK modeling. Section V reports the experimental results on paracetamol and theophylline PBPK systems under both IV and oral dosing scenarios, covering parameter inference and missing term discovery. Finally, Section VI concludes the paper and discusses future directions.

## II. Related Works

### *A. Inverse Modeling in ODE Systems*

Inverse modeling and system identification in ODE systems aim to infer unknown parameters or governing mechanisms from observed system trajectories, and has been widely studied in many areas of science and engineering [18]. Traditional approaches typically rely on numerical solvers coupled with optimization techniques, such as least-squares fitting or gradient-based methods, to estimate model parameters [18]. While effective for small-scale systems, these methods often struggle with high-dimensional, strongly coupled dynamics due to non-convex optimization landscapes and sensitivity to initialization and the computational cost of numerical solvers.

With the rise of machine learning, data-driven approaches have been explored for modeling dynamical systems and addressing inverse problems, including neural ODEs [19], universal differential equations [20], and neural operators [21]. However, purely data-driven methods typically require large amounts of data and may produce physically inconsistent solutions under sparse or noisy observations. Moreover, most existing approaches treat each inverse problem independently, solving a new optimization problem for every system or parameter configuration. For example, Li *et al.* [22] developed gene expression programming to infer governing ODE models from observed data, improving efficiency and accuracy compared to traditional genetic programming approaches. Utkarsh *et al.* [23] proposed a hybrid multi-start optimization strategy that combines particle swarm optimization and L-BFGS to improve convergence in inverse problems for ODEs. Chirigati *et al.* [24] also proposed a Gaussian process-based framework for inverse modeling in dynamical systems, enabling efficient parameter estimation and inference of unobserved components without requiring explicit numerical integration. Despite these advances, such methods generally lack generalization across tasks and often require retraining from scratch, even when the underlying dynamics are similar.

This limitation becomes more pronounced in real-world complex systems, where the underlying dynamics are high-dimensional and only partially observed. A representative example is PBPK modeling. PBPK model construction frequently involves two key steps: (i) determination of drug-specific ADME parameters, and (ii) constructing a system of ODEs that mathematically describes the physiological processes involved in ADME. Conventionally, most parameter values in mechanistic PBPK models are measured through *in vitro* and *in vivo* experiments, while the remaining unknown parameters are empirically estimated through global optimization from clinical data. The innate complexity of PBPK models often presents a scenario where there are many unknown parameters relative to the amount of information available, resulting in poor inference [25]. This has driven the concurrent development of models for first-principles estimation of key parameters [26, 27]. However, while these estimates can be used as model inputs, discrepancies between their simulation outcomes and empirically observed profiles remain an open challenge.

Real-world problems also frequently suffer from incomplete mechanistic knowledge where partial knowledge exists but construction of the complete system of equations purely from first principles is impossible. This creates an impasse in the discovery and mathematical characterization of missing mechanistic terms [28], and modelers are compelled to supplement the system of ODEs with expert knowledge-derived hand-crafted mechanistic terms in order to proceed with conventional parameter estimation. This also remains one of the most significant industry-wide challenges in PK predictions [29].

### *B. Physics-Informed Neural Networks*

PINNs have emerged as a powerful approach for solving ODE-governed problems by embedding known physical laws into the learning objective [30]. Unlike purely data-driven models, PINNs incorporate the governing equations together with initial conditions (ICs) into a unified optimization framework, enabling physically consistent predictions even when observational data is limited or unavailable [31]. This property makes PINNs particularly attractive for scientific and engineering applications where mechanistic knowledge is available but measurements are sparse or noisy [32].

However, most PINN-based approaches formulate inverse modeling as a joint optimization problem, where neural network parameters and unknown physical variables are optimized simultaneously for each task. This often leads to training instability, especially in high-dimensional ODE systems, and makes optimization sensitive to initialization and hyperparameters. More importantly, such formulations remain inherently task-specific and exhibit limited generalization across related systems, requiring retraining from scratch for new configurations.

PINNs have been widely applied to inverse modeling in a variety of dynamical systems, including general physical systems [33, 34], structural mechanics [2], power systems [35], population dynamics [36], and cardiovascular systems [37]. In particular, PINN-based inverse modeling in biomedical ODE systems has attracted increasing attention due to its direct relevance to real-world clinical applications, among which PBPK modeling plays a central role. For instance, Wickramasinghe *et al.* [38] developed a PBPK-iPINN framework to estimate drug- or patient-specific parameters and concentration profiles in PBPK brain compartment models, where a permeability-limited four-compartment structure was considered. Daryakenari *et al.* [28] proposed AI-Aristotle, which combines PINNs with symbolic regression for parameter estimation and missing term discovery, and demonstrated it on a three-compartment pharmacokinetics model and an ultradian glucose-insulin endocrine ODE system. Minadakis [39] applied PINNs to inverse parameter estimation in an eight-compartment PBPK model, identifying tissue partition coefficients from experimental biodistribution measurements. Nevertheless, existing studies remain limited to reduced or organ-specific compartmental models. To the best of our knowledge, PINNs have not yet been demonstrated for whole-body PBPK modeling, which involves substantially higher-dimensional parameter spaces and more complex coupled ODE systems, and yet, is of the most practical relevance for integration with real-world clinical data.

## III. Meta-Inverse Physics-Informed Neural Network

In this section, we introduce MI-PINN, a two-stage framework for inverse learning in high-dimensional ODE systems. MI-PINN first learns the representation across multiple tasks in Stage I, and then performs inverse modeling in Stage II by optimizing only task-specific unknowns while keeping the learned representation fixed. This formulation transforms the original joint optimization problem into a more structured inference process, leading to improved generalization and data efficiency. The overall workflow of MI-PINN is illustrated in Fig. 1.

### *A. Problem Formulation and Task Construction*

ODE systems are widely used to model the dynamics of complex systems. A general nonlinear parameterized system can be expressed as

$$\begin{aligned} &\mathcal{N}\left[u(t;\xi)\right]=\mathcal{S}(t;\xi), && [0,T]\times\Xi \\ &u(0;\xi)=u_0(\xi), && \{t=0\}\times\Xi \\ &u(t;\xi)=u_{\text{Data}}(t;\xi), && [0,T]\times\Xi \end{aligned} \tag{1}$$

where $u(t;\xi)$ denotes the state variables; $\xi\in\Xi$ represents the task-specific system parameters or different missing term formulations used during training; $\mathcal{N}$ is the set of nonlinear differential operators defining the dynamical system; $\mathcal{S}$ is the corresponding source term; $u_0$ is the IC; $u_{\text{Data}}$ represents the sparse observations; $t\in[0,T]$ represents the temporal variable.

For a given set of task-specific parameters $\xi$, the corresponding solution $u(t;\xi)$ can be obtained using conventional numerical solvers. However, inverse tasks such as parameter estimation and missing term discovery typically require repeated evaluations of the governing system under many different parameter configurations. Directly relying on numerical solvers for these evaluations leads to substantial computational cost, especially for large-scale coupled dynamical systems. Therefore, a PINN model provides an efficient alternative for rapid and scalable evaluations.

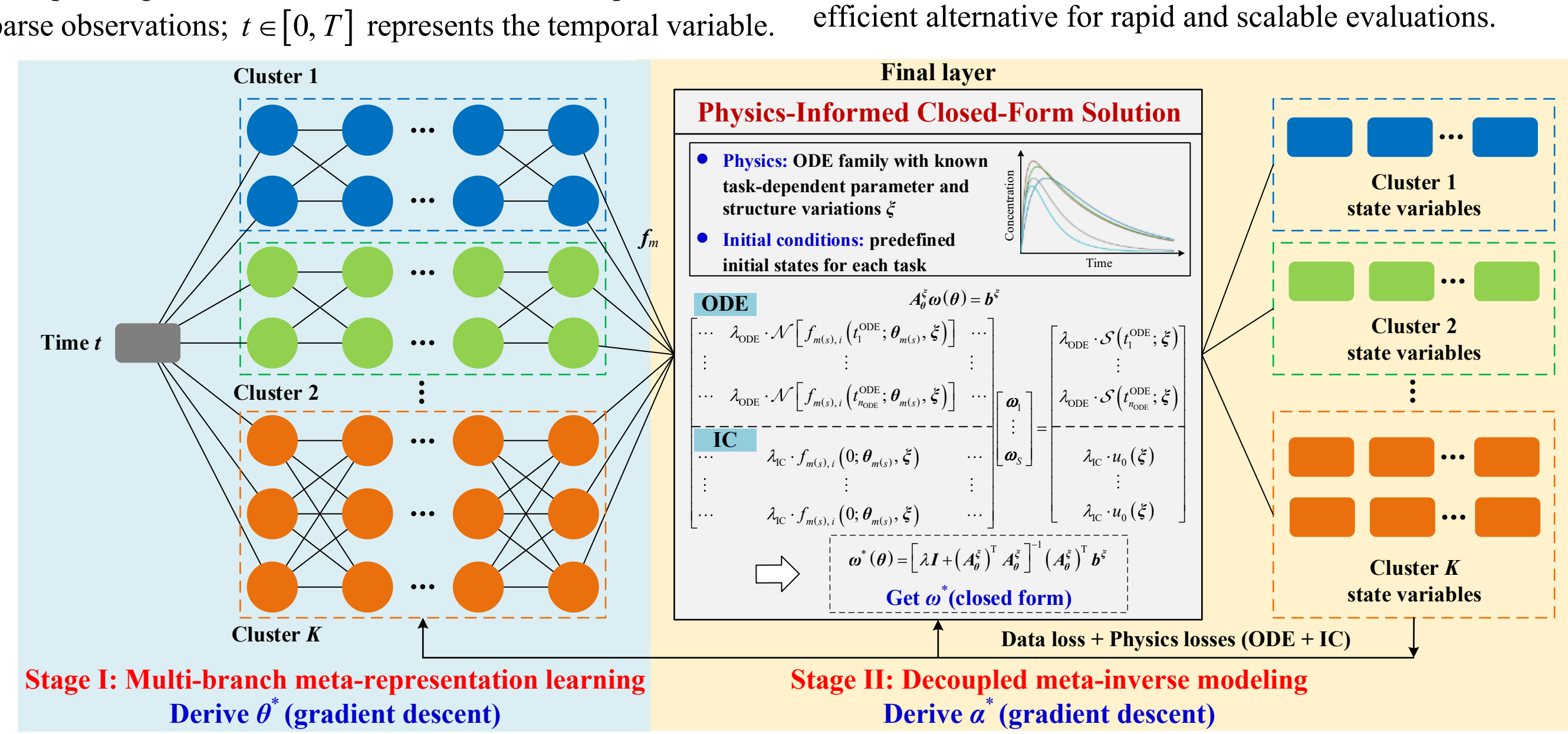

Fig. 1 Overall framework of MI-PINN. In Stage I, MI-PINN employs a multi-branch meta-representation learning scheme to learn the representation parameters $\boldsymbol{\theta}$ across different training task configurations $\xi$ to derive $\boldsymbol{\theta}^*$. The final layer parameters $\boldsymbol{\omega}^*$ are computed via a physics-informed closed-form solution at each optimization step. In Stage II, the learned representation $\boldsymbol{\theta}^*$ is fixed, while the task-specific unknowns $\boldsymbol{\alpha}$ are optimized for inverse modeling to derive $\boldsymbol{\alpha}^*$. The final layer parameters are consistently computed using the same physics-informed closed-form solution, with $\boldsymbol{\alpha}$ replacing $\xi$.

### *B. Physics-Informed Learning Formulation*

For a conventional PINN parameterized by $\boldsymbol{\theta}$, the loss function $\mathcal{L}_{\text{PINN}}$ typically comprises of three components: the IC loss $\mathcal{L}_{\text{IC}}$, the ODE residual loss $\mathcal{L}_{\text{ODE}}$, and, when available, the data-fitting loss $\mathcal{L}_{\text{Data}}$. The overall loss can be expressed as

$$\begin{aligned} &\mathcal{L}_{\text{PINN}}=\lambda_{\text{IC}}\mathcal{L}_{\text{IC}}+\lambda_{\text{ODE}}\mathcal{L}_{\text{ODE}}+\lambda_{\text{Data}}\mathcal{L}_{\text{Data}} \\ &\mathcal{L}_{\text{IC}}=\frac{1}{n_{\text{IC}}}\sum_{i=1}^{n_{\text{IC}}}\left\|u(0;\xi)-u_0(\xi)\right\|_2^2 \\ &\mathcal{L}_{\text{ODE}}=\frac{1}{n_{\text{ODE}}}\sum_{i=1}^{n_{\text{ODE}}}\left\|\mathcal{N}\left[u\left(t_i^{\text{ODE}};\xi\right)\right]-\mathcal{S}\left(t_i^{\text{ODE}};\xi\right)\right\|_2^2 \\ &\mathcal{L}_{\text{Data}}=\frac{1}{n_{\text{Data}}}\sum_{i=1}^{n_{\text{Data}}}\left\|u\left(t_i^{\text{Data}};\xi\right)-u_{\text{Data}}\left(t_i^{\text{Data}};\xi\right)\right\|_2^2 \end{aligned} \tag{2}$$

where $\lambda_{\text{IC}}$, $\lambda_{\text{ODE}}$, and $\lambda_{\text{Data}}$ are the scaling coefficients for the IC, ODE, and data losses, respectively; $n_{\text{IC}}$, $n_{\text{ODE}}$, and $n_{\text{Data}}$ denote the corresponding numbers of sampling points used for each term.

The PINN can be trained using gradient-based optimization methods such as Adam or SGD to minimize the overall loss $\mathcal{L}_{\text{PINN}}$. However, due to the highly non-convex and tightly coupled nature of the physics-informed loss landscape, convergence is often slow and unstable compared with purely data-driven models [40]. These challenges become even more severe in large-scale coupled ODE systems, where different state variables exhibit markedly distinct dynamics, leading to additional difficulties in optimization. Consequently, such limitations make it difficult to directly extend conventional PINNs to real-world inverse problems, thereby motivating the development of a meta-inverse framework such as MI-PINN.

### *C. Two-Stage Meta-Learning Framework for Inverse Modeling*

To address instability and convergence limitations of vanilla PINNs, MI-PINN adopts a two-stage meta-learning framework that first learns a shared physics-aware representation and then performs inverse modeling using the learned representation. The overall framework of MI-PINN is also depicted in Fig. 1.

**Stage I (multi-branch meta-representation learning):** The model is trained across multiple tasks with varying parameter sets or missing term formulations, with the objective of learning a shared physics-aware representation $\boldsymbol{\theta}^*$. The optimization objective is formulated as

$$\left(\boldsymbol{\theta}^*,\boldsymbol{\omega}^*\right)=\arg\min_{\boldsymbol{\theta},\boldsymbol{\omega}}\mathbb{E}_{\xi\sim p(\xi)}\left[\mathcal{L}_{\text{PINN}}\left(\boldsymbol{\theta},\boldsymbol{\omega};\xi\right)\right] \tag{3}$$

where $\boldsymbol{\theta}$ denotes the representation parameters (i.e., all network weights excluding the final layer); $\boldsymbol{\omega}$ represents the final layer parameters. In this stage, an adaptive clustering-based multi-branch representation learning scheme is developed to capture the diverse dynamical behaviors across state variables, while a gradient-free physics-informed closed-form update is developed to stabilize the optimization of the final layer. These two components jointly enable consistent and robust representation learning, as detailed in **Section D**.

**Stage II (decoupled meta-inverse modeling):** For a new target system, the learned representation $\boldsymbol{\theta}^*$ is kept fixed, and MI-PINN solves the inverse problem by optimizing only the task-specific variables using the available observations. The optimization problem is formulated as

$$\boldsymbol{\alpha}^* = \arg\min_{\boldsymbol{\alpha}} \mathcal{L}_{\text{Inverse}}\left(\boldsymbol{\alpha}, \boldsymbol{\omega}\left(\boldsymbol{\alpha}, \boldsymbol{\theta}^*\right); \boldsymbol{\theta}^*\right) \tag{4}$$

where $\boldsymbol{\alpha}$ denotes the task-specific unknowns to be identified during inverse modeling; $\mathcal{L}_{\text{Inverse}}$ denotes the corresponding inverse modeling loss. During this process, the final layer parameters are updated through the same physics-informed closed-form solution, enabling stable and computation-efficient identification even under sparse observations. The details are provided in **Section E**.

The proposed two-stage MI-PINN framework improves both identifiability and training stability in inverse PBPK modeling under sparse observations. In conventional PINNs, high-dimensional network parameters and task-specific variables are optimized jointly, resulting in a large and ill-posed search space that hinders reliable parameter recovery and leads to unstable optimization. In contrast, MI-PINN first learns the representation $\boldsymbol{\theta}^*$ across multiple tasks, which constrains the solution space to a lower-dimensional space, thereby improving identifiability and yielding a better-conditioned optimization problem.

***Proposition 1.*** *Under the learned representation $\boldsymbol{\theta}^*$, the inverse problem in Stage II reduces to identifying the task-specific variables $\boldsymbol{\alpha}$ within a restricted hypothesis space, leading to improved identifiability compared with jointly optimizing all network parameters.*

*Justification.* Since the learned representation $\boldsymbol{\theta}^*$ is fixed, the model output can be expressed as a linear combination of learned basis functions with respect to the final layer parameters. As a result, the effective dimension of the search space is reduced from the full network parameter space to the dimension of $\boldsymbol{\alpha}$, making the inverse problem better conditioned under sparse observations. From an optimization perspective, MI-PINN decouples representation learning from inverse inference. During inverse modeling, only $\boldsymbol{\alpha}$ is optimized, while the final-layer parameters are computed via a physics-informed pseudo-inverse corresponding to a regularized least-squares problem.

***Proposition 2.*** *The closed-form pseudo-inverse solution improves training stability by eliminating gradient-based updates for the final layer.*

*Justification.* The pseudo-inverse provides the optimal solution to the linearized least-squares problem at each iteration, avoiding error accumulation from gradient-based updates and reducing sensitivity to the non-convex loss landscape.

To provide a clearer overview of current approaches, a detailed comparison of existing PINN-based methods for inverse modeling in dynamical systems is summarized in TABLE I. From this comparison, it is evident that most existing works focus on low-dimensional systems and remain inherently task-specific, with limited generalization across related problems. In contrast, MI-PINN is designed for high-dimensional dynamical systems and supports cross-task adaptation under limited observations.

TABLE I
SUMMARY OF PINN-BASED METHODS FOR INVERSE MODELING IN DYNAMICAL SYSTEMS

| Method | System type | Number of observations | System dimension | Inverse task | Generalization |
|---|---|---|---|---|---|
| FedPINN [33] | General physical systems | 1000 | 1-3 (low) | Parameter inference | Task-specific |
| *sf*-PINN [34] | General physical systems | 200-520251 | 1, 3 (low) | Parameter inference | Task-specific |
| Beam PINN [2] | Structural mechanics systems | 5000 | 1, 2 (low) | Parameter inference and force identification | Task-specific |
| Microgrid PINN [35] | Microgrid power systems | - | 2 (low) | Parameter inference | Task-specific |
| Mosquito PINN [36] | Mosquito population dynamics systems | 63-3670 | 3, 10 (low) | Parameter inference | Task-specific |
| Left ventricle PINN [37] | Lumped cardiovascular systems | - | 5 (low) | Parameter inference | Task-specific |
| PBPK-iPINN [38] | PBPK systems | 800 | 4 (low) | Parameter inference | Task-specific |
| AI-Aristotle [28] | PBPK systems | 10-1800 | 3, 6 (low) | Parameter inference and missing term discovery | Task-specific |
| PBPK PINN [39] | PBPK systems | - | 8 (low) | Parameter inference | Task-specific |
| **MI-PINN (Ours)** | **PBPK systems** | **10** | **22, 33 (high)** | **Parameter inference and missing term discovery** | **Cross-task adaptation** |

Note: "-" indicates that the corresponding information is not explicitly reported in the original paper.

### D. Multi-Branch Meta-Representation Learning

To effectively learn multi-scale dynamical behaviors across different compartments, MI-PINN incorporates an adaptive clustering-based multi-branch learning scheme as part of the meta-representation learning process. This is also consistent with current recognition in literature that spectral bias during PINN training can greatly complicate the simultaneous learning of multiple frequency scales. The multi-branch learning scheme is therefore crucial for preventing gradient pathologies due to the multi-scale dynamical state variables.

In MI-PINN, the ODE states across tasks are first grouped into $K$ clusters according to their dynamical similarity. Specifically, for each state variable $u_s$, we collect its trajectory sampled on the collocation time grid and represent it as a vector $\boldsymbol{x}_s$. We then compute the pairwise correlation coefficient $\rho_{ij}$ = corr($\boldsymbol{x}_i$, $\boldsymbol{x}_j$) between state trajectories and define a correlation-based distance metric $d_{ij} = 1 - \rho_{ij}$. Based on the resulting distance matrix $\boldsymbol{D} = [d_{ij}]$, the agglomerative hierarchical clustering is used to partition the ODE states into $K$ groups. In this way, states exhibiting similar temporal profiles are assigned to the same cluster, enabling each branch to specialize in modeling a subset of compartments with coherent dynamics.

Each cluster is then assigned a neural network branch parameterized by $\boldsymbol{\theta}_m$, producing the cluster-specific feature representation $\boldsymbol{f}_m\left(t;\boldsymbol{\theta}_m,\xi\right)$, where $m=1,\cdots,K$. Each state variable $u_s$ is assigned to exactly one cluster, denoted by $m(s)$. The prediction of $u_s$ can be obtained via the linear combination of the features from its own branch as

$$u_s\left(t;\xi\right)=\boldsymbol{f}_{m(s)}\left(t;\boldsymbol{\theta}_{m(s)},\xi\right)^{\mathrm{T}}\boldsymbol{\omega}_s,\ \ s=1,\cdots,S \tag{5}$$

where $S$ denotes the number of ODE state variables; $\boldsymbol{\omega}_s$ represents the final layer parameters associated with state variable $s$. Collecting all state-specific weights yields the final layer parameters $\boldsymbol{\omega}=\left[\boldsymbol{\omega}_1,\cdots,\boldsymbol{\omega}_S\right]$. The branch networks $\left\{\boldsymbol{f}_m\right\}$ provide cluster-specific feature representations, while the state-dependent weights $\left\{\boldsymbol{\omega}_s\right\}$ map these features to the individual state variables. In MI-PINN, the final layer parameters $\boldsymbol{\omega}$ are not updated by gradient descents; instead, they are computed in closed form via a physics-informed pseudo-inverse, as detailed below.

During Stage I, only the representation parameters $\boldsymbol{\theta}=\left\{\boldsymbol{\theta}_1,\cdots,\boldsymbol{\theta}_K\right\}$ are optimized using gradient descent, following the update rule

$$\boldsymbol{\theta}^{*}=\boldsymbol{\theta}^{0}-\eta\sum_{i=0}^{N-1}\nabla_{\boldsymbol{\theta}}\mathcal{L}_{\mathrm{PINN}}\left(\boldsymbol{\theta}^{i},\boldsymbol{\omega}^{*}\left(\boldsymbol{\theta}^{i}\right);\xi\right) \tag{6}$$

where $\eta$ is the learning rate; $N$ is the total number of gradient descent iterations; $\boldsymbol{\omega}^{*}\left(\boldsymbol{\theta}^{i}\right)$ is the closed-form solution of the final layer parameters obtained from the pseudo-inverse at the $i$-th iteration.

In Stage I, the representation parameters $\boldsymbol{\theta}$ (i.e., the parameters prior to the final layer) are learned across multiple tasks with varying parameter sets or missing term formulations $\boldsymbol{\xi}$. To improve the prediction stability, the computation of the final layer parameters $\boldsymbol{\omega}$ is formulated as a physics-informed regularized least-squares problem. This formulation integrates both the feature outputs and the governing system equations, leading to the following optimization problem for $\boldsymbol{\omega}^{*}$

$$\boldsymbol{\omega}^{*}\left(\boldsymbol{\theta}\right)=\arg\min_{\boldsymbol{\omega}}\ \left[\left(\boldsymbol{A}_{\boldsymbol{\theta}}^{\xi}\boldsymbol{\omega}-\boldsymbol{b}^{\xi}\right)^{\mathrm{T}}\left(\boldsymbol{A}_{\boldsymbol{\theta}}^{\xi}\boldsymbol{\omega}-\boldsymbol{b}^{\xi}\right)+\lambda\boldsymbol{\omega}^{\mathrm{T}}\boldsymbol{\omega}\right] \tag{7}$$

where $\boldsymbol{A}_{\boldsymbol{\theta}}^{\xi}$ denotes the physics-informed feature matrix constructed from the learned multi-branch representations $\boldsymbol{f}_m\left(t;\boldsymbol{\theta}_m,\xi\right)$; $\boldsymbol{b}^{\xi}$ represents the physics constraints induced by the governing system; $\lambda$ is the Tikhonov regularization parameter. Since the optimization of the final layer parameters $\boldsymbol{\omega}$ can be formulated as a least-squares problem, the underlying physics can be embedded directly into this formulation, leading to the following system

$$\boldsymbol{A}_{\boldsymbol{\theta}}^{\xi}\boldsymbol{\omega}\left(\boldsymbol{\theta}\right)=\boldsymbol{b}^{\xi}$$

$$\begin{bmatrix}\cdots & \lambda_{\mathrm{ODE}}\cdot\mathcal{N}\left[f_{m(s),i}\left(t_1^{\mathrm{ODE}};\boldsymbol{\theta}_{m(s)},\xi\right)\right] & \cdots\\ \vdots & \vdots & \vdots\\ \cdots & \lambda_{\mathrm{ODE}}\cdot\mathcal{N}\left[f_{m(s),i}\left(t_{n_{\mathrm{ODE}}}^{\mathrm{ODE}};\boldsymbol{\theta}_{m(s)},\xi\right)\right] & \cdots\\ & & \\ \cdots & \lambda_{\mathrm{IC}}\cdot f_{m(s),i}\left(0;\boldsymbol{\theta}_{m(s)},\xi\right) & \cdots\\ \vdots & \vdots & \vdots\\ \cdots & \lambda_{\mathrm{IC}}\cdot f_{m(s),i}\left(0;\boldsymbol{\theta}_{m(s)},\xi\right) & \cdots\end{bmatrix}\begin{bmatrix}\boldsymbol{\omega}_1\\ \vdots\\ \boldsymbol{\omega}_S\end{bmatrix}=\begin{bmatrix}\lambda_{\mathrm{ODE}}\cdot\mathcal{S}\left(t_1^{\mathrm{ODE}};\xi\right)\\ \vdots\\ \lambda_{\mathrm{ODE}}\cdot\mathcal{S}\left(t_{n_{\mathrm{ODE}}}^{\mathrm{ODE}};\xi\right)\\ \\ \lambda_{\mathrm{IC}}\cdot u_0\left(\xi\right)\\ \vdots\\ \lambda_{\mathrm{IC}}\cdot u_0\left(\xi\right)\end{bmatrix} \tag{8}$$

$$i=1,2,\cdots,n_p$$

where $f_{m(s),\,i}$ denotes the $i$-th feature produced by the branch network corresponding to the cluster of state variable $s$; $n_p$ represents the number of features (neurons) in the branch's output layer. It can be seen from Eq. (8) that, once the representation parameters $\boldsymbol{\theta}$ are fixed, the final layer parameters $\boldsymbol{\omega}$ can be directly solved through the physics-informed pseudo-inverse, yielding a closed-form solution when the system is linear with respect to $\boldsymbol{\omega}$ as

$$\boldsymbol{\omega}\left(\boldsymbol{\theta}\right)=\left[\lambda\boldsymbol{I}+\left(\boldsymbol{A}_{\boldsymbol{\theta}}^{\xi}\right)^{\mathrm{T}}\boldsymbol{A}_{\boldsymbol{\theta}}^{\xi}\right]^{-1}\left(\boldsymbol{A}_{\boldsymbol{\theta}}^{\xi}\right)^{\mathrm{T}}\boldsymbol{b}^{\xi}. \tag{9}$$

Even when the system is nonlinear, the lagged-coefficient method [41] can be applied, in which the nonlinear terms are frozen using their values from the previous iteration. This converts the problem into a sequence of linear least-squares subproblems. Consequently, at each iteration, the PINN loss in Eq. (2) can be reformulated in its linearized form with respect to $\boldsymbol{\omega}$ as

$$\begin{aligned}&\mathcal{L}_{\mathrm{PINN}}\left(\boldsymbol{\theta},\boldsymbol{\omega}\left(\boldsymbol{\theta}\right);\xi\right)=\lambda_{\mathrm{IC}}\mathcal{L}_{\mathrm{IC}}+\lambda_{\mathrm{ODE}}\mathcal{L}_{\mathrm{ODE}}+\lambda_{\mathrm{Data}}\mathcal{L}_{\mathrm{Data}}\\ &\mathcal{L}_{\mathrm{IC}}=\frac{1}{n_{\mathrm{IC}}}\sum_{i=1}^{n_{\mathrm{IC}}}\left\|\boldsymbol{f}\left(0;\boldsymbol{\theta},\xi\right)\boldsymbol{\omega}-u_0\left(\xi\right)\right\|_2^2\\ &\mathcal{L}_{\mathrm{ODE}}=\frac{1}{n_{\mathrm{ODE}}}\sum_{i=1}^{n_{\mathrm{ODE}}}\left\|\mathcal{N}\left[\boldsymbol{f}\left(t_i^{\mathrm{ODE}};\boldsymbol{\theta},\xi\right)\right]\boldsymbol{\omega}-\mathcal{S}\left(t_i^{\mathrm{ODE}};\xi\right)\right\|_2^2\\ &\mathcal{L}_{\mathrm{Data}}=\frac{1}{n_{\mathrm{Data}}}\sum_{i=1}^{n_{\mathrm{Data}}}\left\|\boldsymbol{f}\left(t_i^{\mathrm{Data}};\boldsymbol{\theta},\xi\right)\boldsymbol{\omega}-u_{\mathrm{Data}}\left(t_i^{\mathrm{Data}};\xi\right)\right\|_2^2\end{aligned} \tag{10}$$

In practical applications, MI-PINN leverages Eq. (10) to efficiently process prior tasks with different configurations $\boldsymbol{\xi}$, each corresponding to a distinct task instance or system setting. By repeatedly solving the linearized system and updating the final layer under physics constraints, the training process becomes more stable and converges significantly faster. This iterative closed-form refinement greatly improves computational efficiency and predictive accuracy in meta-learning scenarios involving high-dimensional coupled ODE systems.

### *E. Decoupled Meta-Inverse Modeling*

In Stage II, the pretrained representation parameters $\boldsymbol{\theta}^*$, which encapsulate the knowledge learned from multiple tasks $\xi$, are kept fixed. The model then identifies the task-specific variables for a new target system by optimizing only the corresponding inverse variables, including physical parameters and neural parameters associated with missing term discovery, as formulated in

$$\begin{aligned}\boldsymbol{\alpha}^* &= \boldsymbol{\alpha}^0 - \sum_{i=0}^{N-1} \eta \nabla_{\alpha} \mathcal{L}_{\text{Inverse}}\left(\boldsymbol{\alpha}^i, \boldsymbol{\omega}^*\left(\boldsymbol{\alpha}^i, \boldsymbol{\theta}^*\right); \boldsymbol{\theta}^*\right) \\ \boldsymbol{\omega}^*\left(\boldsymbol{\alpha}, \boldsymbol{\theta}^*\right) &= \arg\min_{\boldsymbol{\omega}} \left[\left(\boldsymbol{A}_{\boldsymbol{\theta}^*}^{\boldsymbol{\alpha}} \boldsymbol{\omega} - \boldsymbol{b}^{\boldsymbol{\alpha}}\right)^{\mathrm{T}} \left(\boldsymbol{A}_{\boldsymbol{\theta}^*}^{\boldsymbol{\alpha}} \boldsymbol{\omega} - \boldsymbol{b}^{\boldsymbol{\alpha}}\right) + \lambda \boldsymbol{\omega}^{\mathrm{T}} \boldsymbol{\omega}\right]\end{aligned} \tag{11}$$

where $\boldsymbol{\alpha}$ denotes the task-specific unknown variables to be identified during inverse modeling, such as physical parameters and neural parameters associated with missing term discovery. Since the pretrained representation parameters $\boldsymbol{\theta}^*$ are kept fixed, the final layer parameters $\boldsymbol{\omega}$ can still be obtained via the physics-informed pseudo-inverse as defined in Eqs. (8) and (9), but now with $\boldsymbol{\alpha}$ replacing the previously known task-specific configurations $\xi$. The corresponding loss function for inverse modeling is formulated as

$$\begin{aligned}&\mathcal{L}_{\text{Inverse}}\left(\boldsymbol{\alpha}, \boldsymbol{\omega}\left(\boldsymbol{\alpha}, \boldsymbol{\theta}^*\right); \boldsymbol{\theta}^*\right) = \\ &\lambda_{\text{ODE}} \frac{1}{n_{\text{ODE}}} \sum_{i=1}^{n_{\text{ODE}}} \left\| \mathcal{N}\left[\boldsymbol{f}\left(t_i^{\text{ODE}}; \boldsymbol{\alpha}, \boldsymbol{\theta}^*\right)\right] \boldsymbol{\omega} - \mathcal{S}\left(t_i^{\text{ODE}}; \boldsymbol{\alpha}\right) \right\|_2^2 \\ &+ \lambda_{\text{Data}} \frac{1}{n_{\text{Obs}}} \sum_{i=1}^{n_{\text{Obs}}} \left\| \boldsymbol{f}\left(t_i^{\text{Obs}}; \boldsymbol{\alpha}, \boldsymbol{\theta}^*\right) \boldsymbol{\omega} - u_{\text{Obs}}\left(t_i^{\text{Obs}}\right) \right\|_2^2\end{aligned} \tag{12}$$

where $n_{\text{Obs}}$ denotes the number of observation points used for inverse modeling; $u_{\text{Obs}}$ represents the corresponding ground truth observation data. During this stage, gradient-based optimizers such as SGD or Adam are employed to update $\boldsymbol{\alpha}$ in Eq. (12) using the available observation data, while $\boldsymbol{\theta}^*$ remains fixed and the final layer parameters $\boldsymbol{\omega}$ are iteratively computed via the physics-informed pseudo-inverse based on the current estimate of $\boldsymbol{\alpha}$. This meta-informed inverse modeling procedure enables MI-PINN to recover unknown physical parameters and model missing terms in large-scale dynamical systems, thereby providing a stable and reliable framework for the high-dimensional ODE system identification.

## IV. MI-PINN for Inverse PBPK Modeling

### *A. PBPK Model*

To demonstrate the capability of MI-PINN on a realistic nonlinear, high-dimensional, and multi-scale ODE system, we choose to utilize the whole-body PBPK modeling framework described here. The whole-body PBPK model is used to describe the ADME behaviour of drugs as they enter the body and consists of multiple organ compartments, each characterized by their own distinct volumes, blood flow and partition coefficients, collectively forming a comprehensive whole-body system. These organs include the brain, lungs, heart, stomach, intestine, spleen, pancreas, adipose tissue, skin, muscle, bones, liver, and kidney. The liver and kidney serve as the primary sites for systemic drug elimination, while intestinal and hepatic first-pass metabolism account for pre-systemic clearance. Arterial and venous blood compartments connect these organs, with drug distribution throughout the body determined by organ-specific blood flow rates and tissue partition coefficients. Tissue-to-plasma partition coefficients are calculated using the method described by [42], which governs the extent of drug partitioning across different tissues. The overall model structure is illustrated in Fig. 2.

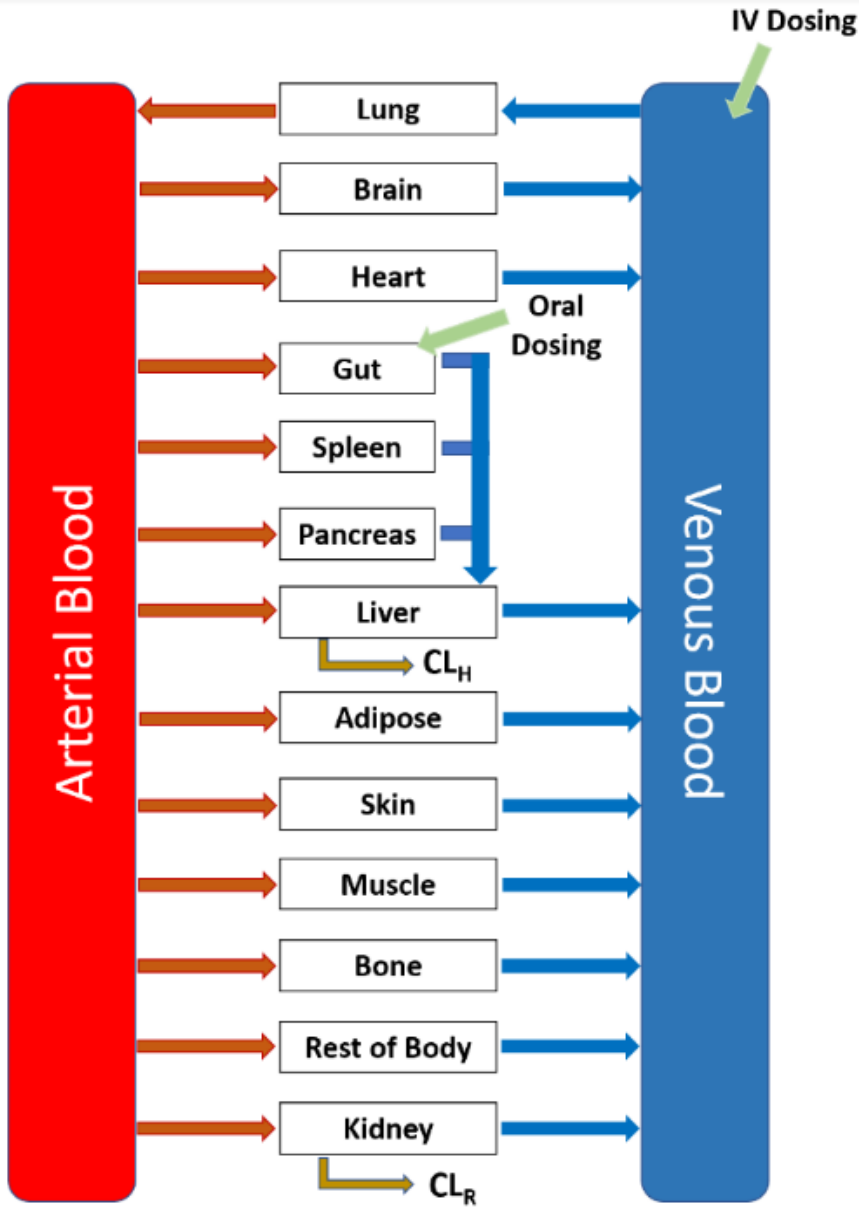

Fig. 2 A schematic of a classical PBPK model.

The transfer of drugs between all compartments follows a perfusion-limited distribution which assumes the rate-determining steps of drug distribution to be governed by tissue blood flow rather than permeability factors. This assumption is commonly used and valid when working with drugs with high permeability. Thus, the rate of change of drug in each organ compartment can be described by the following mass balance equation

$$\frac{dC_{\text{Organ}}}{dt} = \frac{Q_{\text{Organ}}\left(C_{\text{ArterialBlood}} - C_{\text{Organ}} R_b / Kp_{\text{Organ}}\right)}{V_{\text{Organ}}} \tag{13}$$

where $C_{\text{Organ}}$ denotes the drug concentration in the organ compartment; $Q_{\text{Organ}}$ is the organ blood flow rate; $C_{\text{ArterialBlood}}$ is the drug concentration in the arterial blood compartment; $R_b$ represents the blood-to-plasma ratio of the drug; $Kp_{\text{Organ}}$ denotes the tissue-to-plasma partition coefficient of the drug; $V_{\text{Organ}}$ is the organ volume. The equation for lungs is identical but with the substitution of $C_{\text{ArterialBlood}}$ with venous blood drug concentration $C_{\text{VenousBlood}}$ due to reverse flow directionality from the venous blood to the lungs and back to arterial blood.

The gut compartments become relevant for oral administration by accounting for drug dissolution and other absorption-related processes. The absorption rate of drugs from the intestinal lumen to intracellular can be defined by the absorption rate constant $k_a$ using the following equation

$$k_a = \frac{2P_{\text{eff, man}}}{R} \tag{14}$$

where $P_{\text{eff, man}}$ represents the effective permeability in man and is computed from *in vitro* apparent permeability values using the relationships established by [43], and $R$ represents the radius of the intestine.

Drugs that are absorbed in the intestinal tract are subjected to first-pass effect when they enter the liver via the portal vein before being distributed to the rest of the body by blood flows. Drug clearance in the liver can involve multiple CYP, UGT and SULT enzymes depending on the drug and can be described using the following equation

$$\frac{dC_{\text{Liver}}}{dt} = \Big[ Q_{\text{Liver}} \left( C_{\text{ArterialBlood}} - C_{\text{Liver}} R_b / Kp_{\text{Liver}} \right) - \sum_{i=1}^{N} \frac{V_{\max,i} \times C_{\text{Liver}} \times f_{u,b} \times MPPGL \times W_{\text{Liver}}}{K_{m,i} \times f_{u,\text{mic}} + C_{\text{Liver}} \times f_{u,b}} \Big] \Big/ V_{\text{Liver}} \quad (15)$$

where $V_{\max,i}$ denotes the maximal reaction rate for the $i$-th enzyme in drug metabolism; $f_{u,b}$ is the unbound fraction of the drug; $MPPGL$ represents the microsomal protein per gram of liver (replaced with $CPPGL$ representing cytosolic protein per gram of liver for SULT enzymes); $W_{\text{Liver}}$ is the liver weight; $K_{m,i}$ is the drug concentration which provides half of its maximal reaction rate for the $i$-th enzyme. The *in vitro* $V_{\max}$ is scaled to whole liver level using *in vitro*-to-*in vivo* extrapolation (*IVIVE*) with the *MPPGL* and $W_{\text{Liver}}$ terms.

Drugs can also be renally cleared and are assumed to follow a simple linear elimination kinetics in the kidney compartment, defined by the following equation

$$\frac{dC_{\text{Kidney}}}{dt} = \frac{Q_{\text{Kidney}} \left( C_{\text{ArterialBlood}} - C_{\text{Kidney}} R_b / Kp_{\text{Kidney}} \right) - CL_R}{V_{\text{Kidney}}} \quad (16)$$

where $CL_R$ represents the renal clearance of the drug.

### B. *Inverse PBPK Modeling Formulation*

Based on the general dynamical system formulation introduced in Section III.A, the whole-body PBPK system described in Section IV.A can be viewed as a specific instance of a parameterized dynamical system. Hence, the PBPK model is a representative high-dimensional system of coupled nonlinear ODEs, where the state variables correspond to drug or metabolite concentrations across multiple physiological compartments. In this context, the state variable $u(t;\xi)$ in Eq. (1) corresponds to drug or metabolite concentrations across physiological compartments, while the task-specific parameters $\xi$ denote the input configurations that define PBPK tasks, where different configurations correspond to variations in key kinetic parameters or different formulations of missing mechanistic terms, thereby inducing a family of related PBPK systems.

During Stage I, MI-PINN is trained across this family of PBPK tasks induced by varying configurations $\xi$. Through this process, the model learns the representation parameters $\boldsymbol{\theta}^*$ that capture common dynamical structures across multi-scale PBPK systems.

For a new target PBPK system, the inverse problem aims to infer unknown system components from sparse and noisy clinical observations. In this work, we consider two types of inverse tasks based on the same observed concentration-time data: parameter inference, where unknown kinetic parameters are estimated, and missing term discovery, where unknown mechanistic components in the governing equations are reconstructed.

In Stage II, the pretrained representation $\boldsymbol{\theta}^*$ is fixed during inverse modeling, and only the task-specific variables $\boldsymbol{\alpha}$ in Eq. (11) are optimized. In the context of PBPK modeling, $\boldsymbol{\alpha}$ may correspond to unknown physiological or kinetic parameters, or to the parameters of auxiliary neural networks introduced to approximate missing mechanisms across compartments as illustrated in Fig. 3. Meanwhile, the final layer parameters $\boldsymbol{\omega}$ are computed via the physics-informed pseudo-inverse in Eqs. (8) and (9) based on the current estimate of $\boldsymbol{\alpha}$. This two-stage formulation effectively decouples representation learning and inverse modeling and reduces the complexity of the optimization problem, thereby providing a unified framework of parameter inference and missing term discovery in PBPK systems.

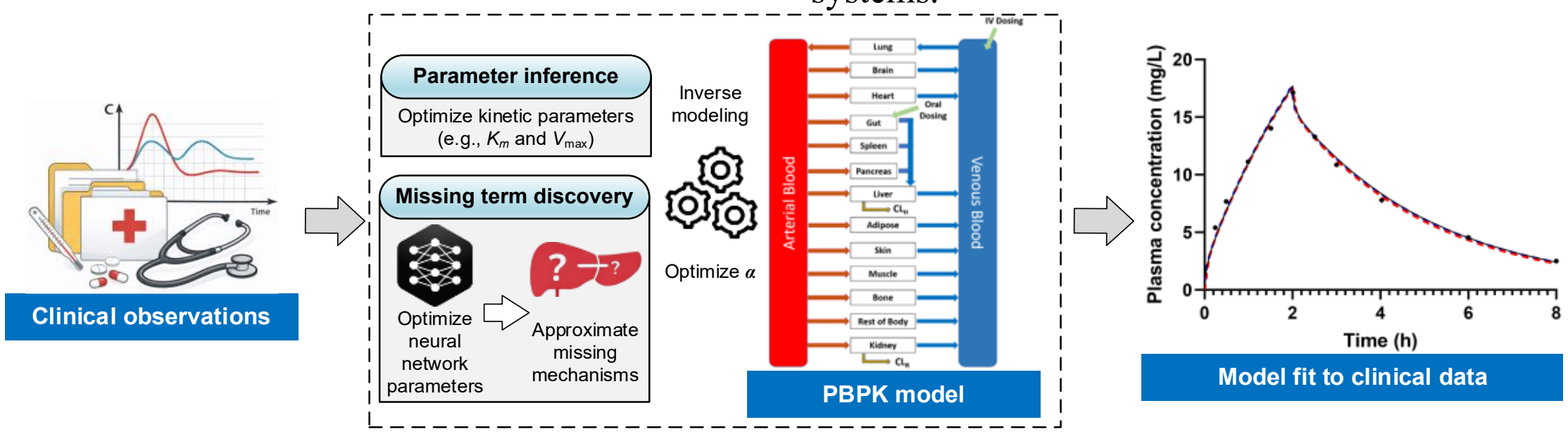

Fig. 3 Illustration of parameter inference and missing term discovery in PBPK modeling.

## V. Experimental Study

The MI-PINN framework is extensively tested with multiple scenarios. To establish a working proof-of-concept and assess its performance in parameter inference and missing mechanism discovery, paracetamol is used as a case-study. Paracetamol is chosen as a first demonstration because of its well-characterized PK and well-described clearance pathways. The drug parameters for the paracetamol PBPK model are reported in Supplementary Table S1. The paracetamol model is tested under two different dosing administration routes: the IV dosing case (22 ODE-system) and oral case (33 ODE-system). A second model using theophylline as a case-study is constructed to assess the robustness of the core MI-PINN framework in providing accurate parameter inference when applied to a different drug with different parameter sets. While the paracetamol model is tested and validated using aggregate data, in the case of theophylline, multiple individual clinical data is used to assess the model's ability to handle noisy data arising from inter-individual variability. The theophylline model drug parameters are reported in Supplementary Table S2. The full set of ODEs for the IV and oral scenarios are reported in Supplementary Tables S3 and S4, respectively, while lists of compartment and system parameters used in the modeling are reported in Supplementary Tables S5 and S6, respectively. The summary of ODE structure and coupling in PBPK models is shown in TABLE II.

For the IV and oral dosing case, the number of clusters is determined to be two and three respectively. Each branch network adopts a two-layer architecture with 256 neurons per layer and sine activation functions, denoted as 1-256(sin)-256(sin). The outputs of all branch networks are concatenated and mapped to the final state variables through a final layer, whose parameters are computed via the proposed physics-informed pseudo-inverse. All experiments are trained using the Adam optimizer with an initial learning rate of 0.05. The number of temporal input points is set to 200. For missing term discovery, an auxiliary neural network with architecture 1-3(swish)-1(linear) is employed to approximate the unknown mechanistic components.

TABLE II
SUMMARY OF ODE STRUCTURE AND COUPLING IN PBPK MODELS

| Model | Number of ODEs | ODE coupling structure |
|---|---|---|
| IV paracetamol PBPK | 22 | 20 linear, 2 nonlinear |
| Oral paracetamol PBPK | 33 | 21 linear, 12 nonlinear |
| IV theophylline PBPK | 22 | 20 linear, 2 nonlinear |

### *A. IV Paracetamol PBPK Model for Parameter Inference*

To evaluate the effectiveness of the proposed MI-PINN, we first consider an IV paracetamol dosing case described by a PBPK system consisting of 22 ODEs, including 20 linear and 2 nonlinear equations (Supplementary Table S3). The system can be generally written as

$$\dot{\boldsymbol{x}}(t) = \boldsymbol{f}\left(\boldsymbol{x}(t), \boldsymbol{p}, t\right), \quad \boldsymbol{x}(0) = \boldsymbol{x}_0 \tag{17}$$

where $\boldsymbol{x}(t)$ denotes the vector of state variables (i.e., drug concentrations across compartments); $\boldsymbol{p}$ denotes the PBPK model parameters; $\boldsymbol{f}(\cdot)$ defines the underlying PBPK dynamics; $\boldsymbol{x}_0$ represents the initial conditions.

In this modeling exercise, we intentionally mask the parameter value for the $V_{\max}$ and $K_m$ of UDP-glucuronosyltransferase (UGT) 2B15 enzyme metabolism of paracetamol to evaluate if MI-PINN can predict the correct values. UGT2B15 enzyme kinetics is chosen to be masked as glucuronidation by UGTs accounts for majority of paracetamol's metabolic pathway, with the 2B15 isoform possessing the highest *in vitro* clearance value amongst the UGTs [44]. The PK profile for paracetamol is therefore expected to be sensitive to UGT2B15 kinetics making them a suitable candidate for testing parameter inference. The observed output data used for the fitting exercise is the aggregated venous blood concentration-time profile (consisting of 10 clinical observation points) for paracetamol after a two-hour 20 mg/kg infusion dose from a clinical study [45].

As a baseline, the whole-body PBPK model is separately constructed in MATLAB® version R2023b (23.2.0.2380103) (MathWorks, Natick, USA) using the SimBiology® module (version 23.2). We then perform parameter inference simultaneously for $V_{\max}$ and $K_m$ of UGT2B15 metabolism kinetics (denoted as $K_{m,\,P}$ and $V_{\max,\,P}$ for paracetamol) using MATLAB's *lsqcurvefit* function which employs a trust region reflective algorithm. The algorithm iteratively fits model parameters by locally approximating the objective function within a dynamically adjusted trust region and enforcing parameter bounds via reflection, thereby allowing convergence for such constrained nonlinear least-squares problems [46].

Parameter inference is separately performed with MI-PINN. In Stage I, MI-PINN is trained according to Eq. (10) to learn the representation parameters $\boldsymbol{\theta}^*$ via meta-learning over 20 tasks with different parameter configurations of the two enzyme kinetic parameters $K_{m,\,P}$ and $V_{\max,\,P}$ (i.e., $\boldsymbol{\xi}$ in Eq. (10)). In Stage II, the pretrained representation $\boldsymbol{\theta}^*$ is kept fixed, and the kinetic parameters $K_{m,\,P}$ and $V_{\max,\,P}$ (i.e., $\boldsymbol{\alpha}$ in Eq. (11)) are inferred by solving Eqs. (11) and (12) using the same clinical data [45] used with MATLAB's trust region reflective algorithm. The prediction results for all 22 state variables obtained via parameter inference are presented in Fig. 4, while the corresponding physiological interpretations of State 1-State 22 are provided in TABLE III. The inferred parameters obtained by MI-PINN and the MATLAB trust region reflective algorithm are summarized in TABLE IV.

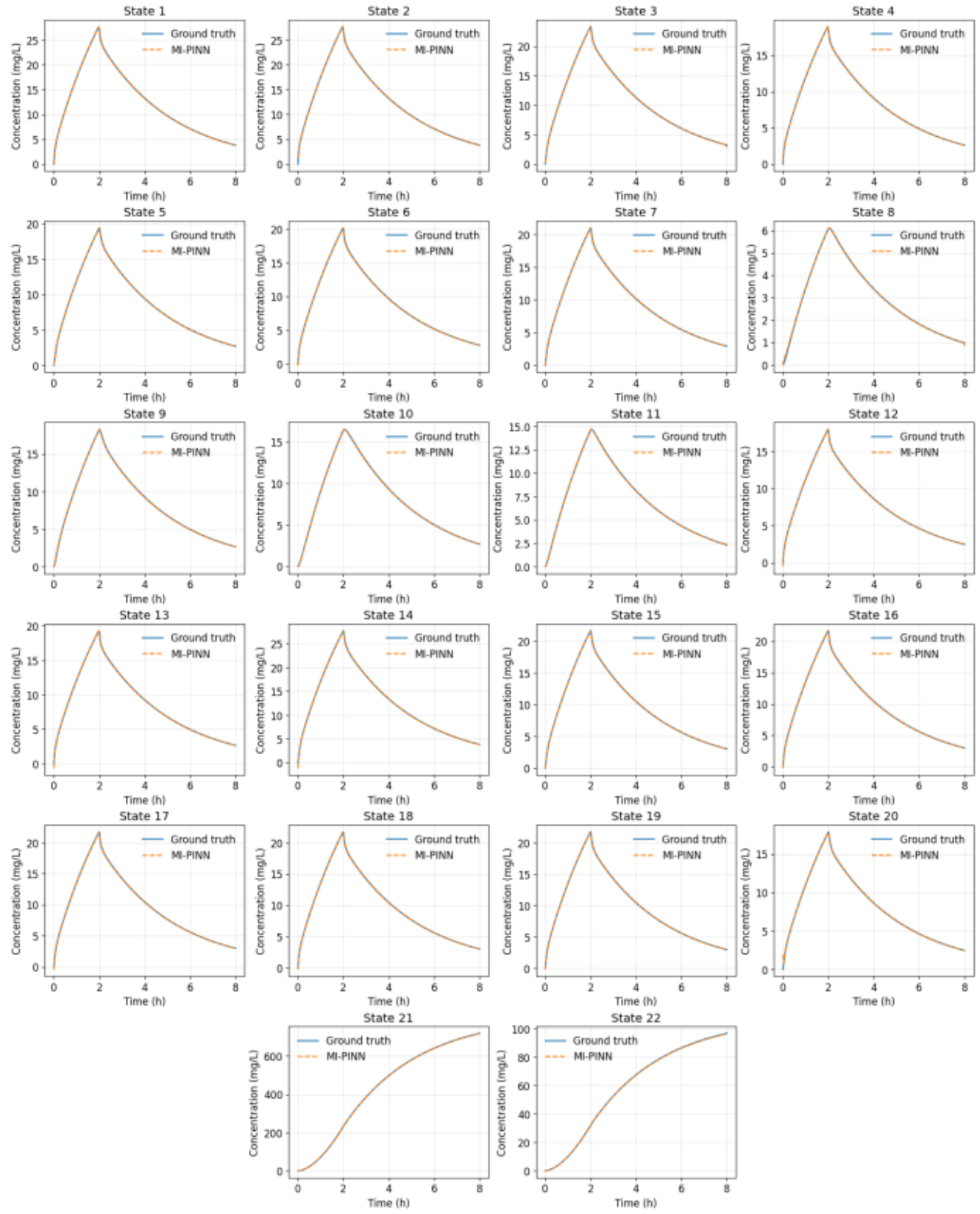

Fig. 4 Prediction results of all 22 state variables obtained via parameter inference for the IV paracetamol PBPK model.

It can be seen from Fig. 4 that the predictions produced by the proposed MI-PINN exhibit excellent agreement with the reference results. The mean squared error (MSE) across all 22 state variables is only 0.96, demonstrating the strong predictive capability of the model. As summarized in TABLE IV, while the ratio of $V_{\max,\,P}$ to $K_{m,\,P}$ (the ratio represents the catalytic efficiency of UGT2B15) inferred by both the trust region reflective algorithm and MI-PINN are similar to the actual values, both $V_{\max,\,P}$ and $K_{m,\,P}$ predictions are significantly more accurate with MI-PINN versus the MATLAB algorithm, even when trained using a single aggregated venous blood concentration-time profile. These results indicate that MI-PINN can reliably recover key kinetic parameters under sparse clinical data.

TABLE III
DEFINITIONS OF STATE VARIABLES FOR STATE 1-STATE 22

| State | Description | State | Description |
|---|---|---|---|
| State 1 | Arterial blood (drug) | State 12 | Rest of body (drug) |
| State 2 | Venous blood (drug) | State 13 | Kidney (drug) |
| State 3 | Brain (drug) | State 14 | Portal vein (drug) |
| State 4 | Lung (drug) | State 15 | Stomach (drug) |
| State 5 | Heart (drug) | State 16 | Duodenum wall (drug) |
| State 6 | Spleen (drug) | State 17 | Jejunum wall (drug) |
| State 7 | Pancreas (drug) | State 18 | Ileum wall (drug) |
| State 8 | Adipose (drug) | State 19 | Large intestine wall (drug) |
| State 9 | Skin (drug) | State 20 | Liver (drug) |
| State 10 | Muscle (drug) | State 21 | Liver (metabolite) |
| State 11 | Bone (drug) | State 22 | Urine (excreted drug) |

TABLE IV
INFERRED KINETIC PARAMETERS FOR THE IV PARACETAMOL PBPK MODEL

| Model | $K_{m,P}$ (μmol/L) | $V_{\max,P}$ (μmol/(h·mg)) | Ratio of $V_{\max,P}$ to $K_{m,P}$ |
|---|---|---|---|
| Ground truth [45] | 23000 | 1.80 | $7.83 \times 10^{-5}$ |
| Trust region reflective algorithm | 100,000 | 8.86 | $8.86 \times 10^{-5}$ |
| **MI-PINN** | **27,727** | **2.18** | $\mathbf{7.86 \times 10^{-5}}$ |

### B. Oral Paracetamol PBPK Model for Parameter Inference

To further validate the practical scalability of MI-PINN, we consider an oral dosing case involving the ingestion of a 1000 mg paracetamol tablet [45] using a more complex PBPK system with 33 ODEs, comprising 21 linear and 12 nonlinear equations, from the addition of gut compartments and absorption processes (Supplementary Table S4). The system can be written in the same general form as Eq. (17).

MI-PINN is applied to the oral dosing case under the same two-stage training and inverse inference settings as described in Section V.A. Specifically, the representation parameters $\boldsymbol{\theta}^*$ are learned via meta-training on the same 20 parameter sets of $K_{m,P}$ and $V_{\max,P}$, while all inverse inference settings, including the clinical observation data and the kinetic parameters to be inferred, remain unchanged. The prediction results for all 33 state variables obtained via parameter inference are presented in Fig. 5, while the physiological interpretations of State 1-State 33 are summarized in TABLE V. The inferred parameters obtained by MI-PINN and MATLAB are shown in TABLE VI.

As shown in Fig. 5, even for the more complex oral paracetamol PBPK model consisting of 33 ODEs (12 nonlinear), the proposed MI-PINN achieves highly accurate predictions. Furthermore, as summarized in TABLE VI, the ratio between $V_{\max,P}$ and $K_{m,P}$ inferred by MI-PINN remains close to the ground truth. In contrast, the MATLAB-based trust-region reflective algorithm exhibits degraded parameter consistency in the higher-dimensional PBPK system, demonstrating the robustness and effectiveness of the proposed approach for inverse parameter inference despite increased model complexity.

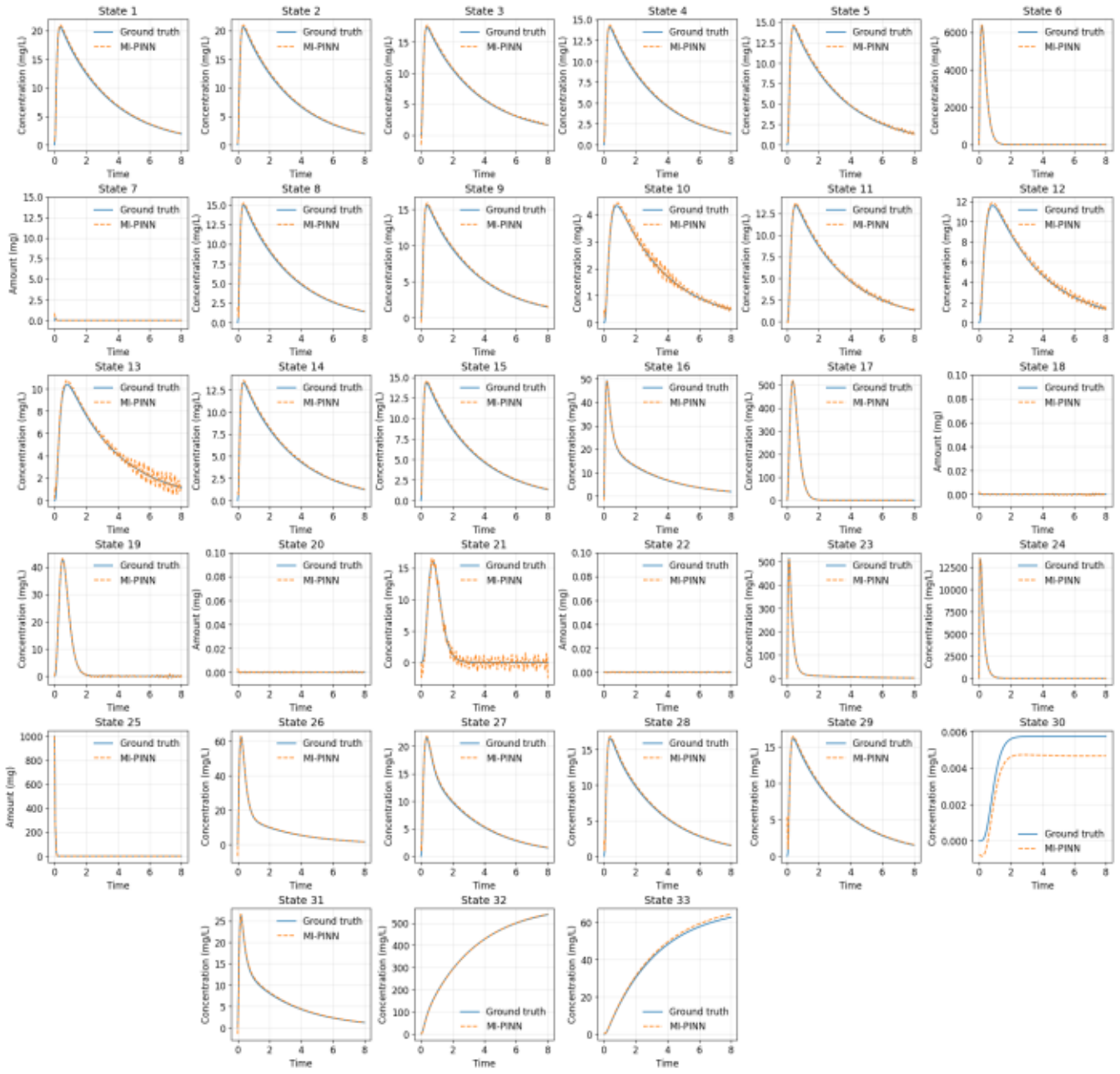

Fig. 5 Prediction results of all 33 state variables obtained via parameter inference for the oral paracetamol PBPK model.

TABLE V
DEFINITIONS OF STATE VARIABLES FOR STATE 1-STATE 33

| State | Description | State | Description |
|---|---|---|---|
| State 1 | Arterial blood (drug) | State 18 | Jejunum lumen (solid drug) |
| State 2 | Venous blood (drug) | State 19 | Ileum lumen (dissolved drug) |
| State 3 | Brain (drug) | State 20 | Ileum lumen (solid drug) |
| State 4 | Lung (drug) | State 21 | Large intestine lumen (dissolved drug) |
| State 5 | Heart (drug) | State 22 | Large intestine lumen (solid drug) |
| State 6 | Duodenum lumen (dissolved drug) | State 23 | Stomach (drug) |
| State 7 | Duodenum lumen (solid drug) | State 24 | Stomach fluid (dissolved drug) |
| State 8 | Spleen (drug) | State 25 | Stomach fluid (solid drug) |
| State 9 | Pancreas (drug) | State 26 | Duodenum wall (drug) |
| State 10 | Adipose (drug) | State 27 | Jejunum wall (drug) |
| State 11 | Skin (drug) | State 28 | Ileum wall (drug) |
| State 12 | Muscle (drug) | State 29 | Large intestine wall (drug) |
| State 13 | Bone (drug) | State 30 | Feces (excreted drug) |
| State 14 | Rest of body (drug) | State 31 | Liver (drug) |
| State 15 | Kidney (drug) | State 32 | Liver (metabolite) |
| State 16 | Portal vein (drug) | State 33 | Urine (excreted drug) |
| State 17 | Jejunum lumen (dissolved drug) | - | - |

TABLE VI
INFERRED KINETIC PARAMETERS FOR THE ORAL PARACETAMOL PBPK MODEL

| Model | $K_{m,P}$ (μmol/L) | $V_{\max,P}$ (μmol/(h·mg)) | Ratio of $V_{\max,P}$ to $K_{m,P}$ |
|---|---|---|---|
| Ground truth [45] | 23,000 | 1.80 | $7.83 \times 10^{-5}$ |
| Trust region reflective algorithm | 16,310 | 3.74 | $2.29 \times 10^{-4}$ |
| **MI-PINN** | **18,053** | **1.32** | $\mathbf{7.31 \times 10^{-5}}$ |

### C. IV Paracetamol PBPK Model for Neural-Symbolic Discovery of Unknown Mechanisms

To further demonstrate the versatility of MI-PINN, we consider the potential for missing mechanism discovery based on the same IV paracetamol PBPK system. In contrast to the parameter inference setting in Section V.A, we intentionally mask the entire UGT2B15 Michaelis-Menten equation including the *IVIVE* terms in the ODEs corresponding to concentration of paracetamol and its metabolite in the liver, and evaluate MI-PINN's ability to handle such under-specified systems. Specifically, the entire UGT2B15 metabolic equation, as formulated in Eq. (18), is now treated as unknown

$$V_{\max,P}/(K_{m,P} \times f_{u,\text{mic}} + C_{\text{Liver}}/M_W \times f_{u,b}) \times MPPGL \times W_{\text{Liver}} \times C_{\text{Liver}} \times f_{u,b} \times R_b / \left(V_{\text{Liver}} \times Kp_{\text{Liver}}\right). \tag{18}$$

where $f_{u,\text{mic}}$ denotes the fraction unbound of paracetamol in the microsomes (included to improve *IVIVE* of clearance); $C_{\text{Liver}}$ represents the paracetamol concentration in the liver; $f_{u,b}$ is the fraction unbound in blood; $M_W$ denotes the molecular weight of paracetamol; $R_b$ represents the blood-to-plasma ratio; $Kp_{\text{Liver}}$ is the tissue partition coefficient between liver and plasma; $V_{\text{Liver}}$ denotes the liver volume.

During meta-training, 20 different candidate formulations of the missing term (i.e., $\boldsymbol{\xi}$ in Eq. (10)) are used to learn the optimized representation parameters $\boldsymbol{\theta}^*$ via Eq. (10). With the pretrained representation $\boldsymbol{\theta}^*$ fixed, the unknown term is approximated by a neural network whose parameters (i.e., $\boldsymbol{\alpha}$ in Eq. (11)) are optimized using the same clinical data as in Section V.A by solving Eqs. (11) and (12). The prediction results for all 22 state variables obtained through this approach are presented in Fig. 6. The physiological interpretations of State 1-State 22 follow the same definitions as those summarized in TABLE III. The approximated mechanistic term's trajectory over time is illustrated in Fig. 7.

As shown in Fig. 6, even in the presence of a missing term, the proposed MI-PINN achieves highly accurate predictions across all 22 state variables, with an overall MSE of approximately 4.73. This demonstrates that the learned feature representations and physics-informed pseudo-inverse formulation enable MI-PINN to maintain predictive stability and generalization, even under incomplete mechanistic knowledge. Furthermore, as illustrated in Fig. 7, the reconstructed missing mechanistic term closely matches the ground-truth, confirming MI-PINN's capability to identify hidden physical relationships and approximate unknown governing dynamics directly from limited and noisy observational data. These results highlight the potential of MI-PINN as a powerful framework for scientific discovery in complex biophysical systems.

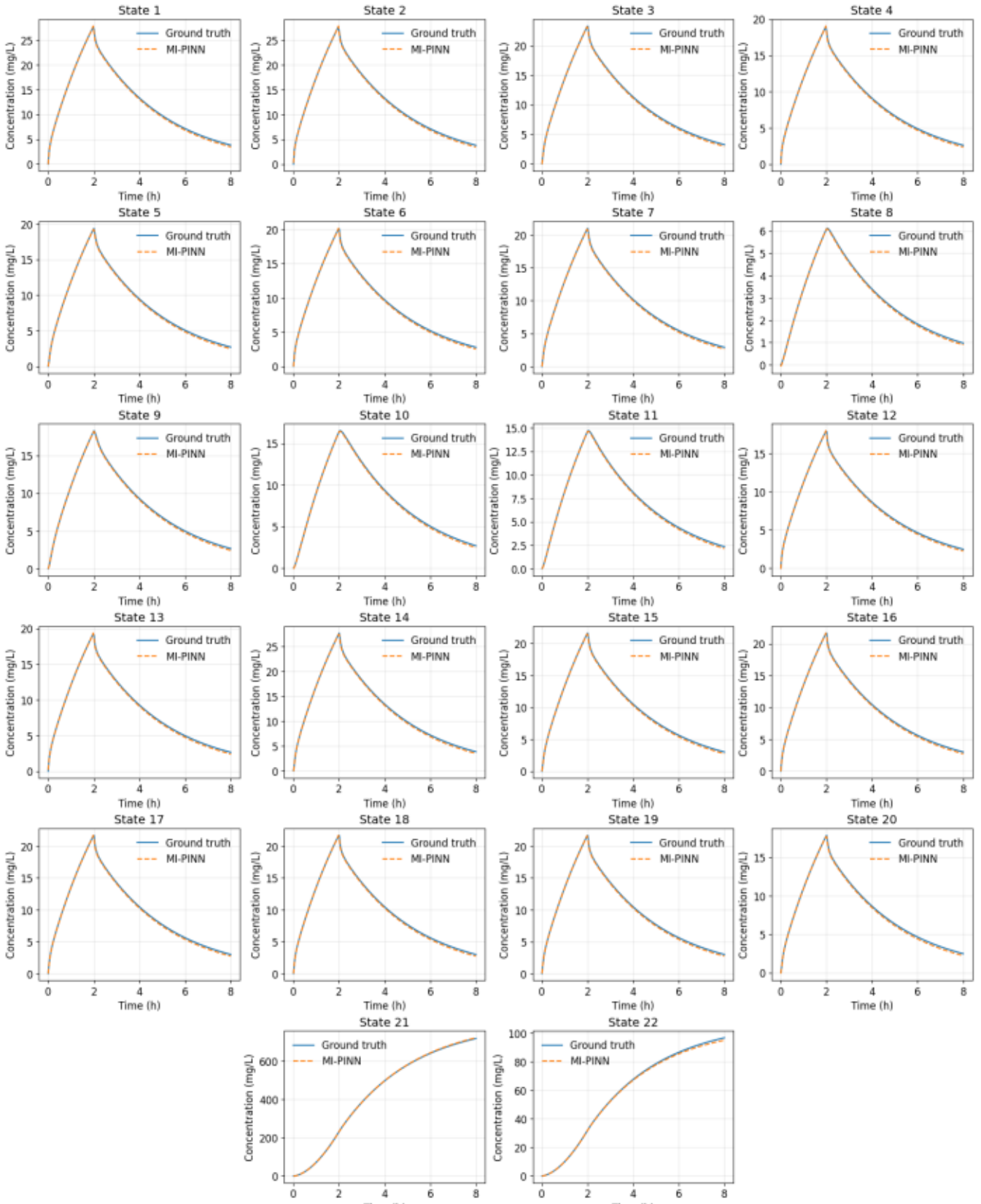

Fig. 6 Prediction results of all 22 state variables obtained via missing term discovery for the IV paracetamol PBPK model.

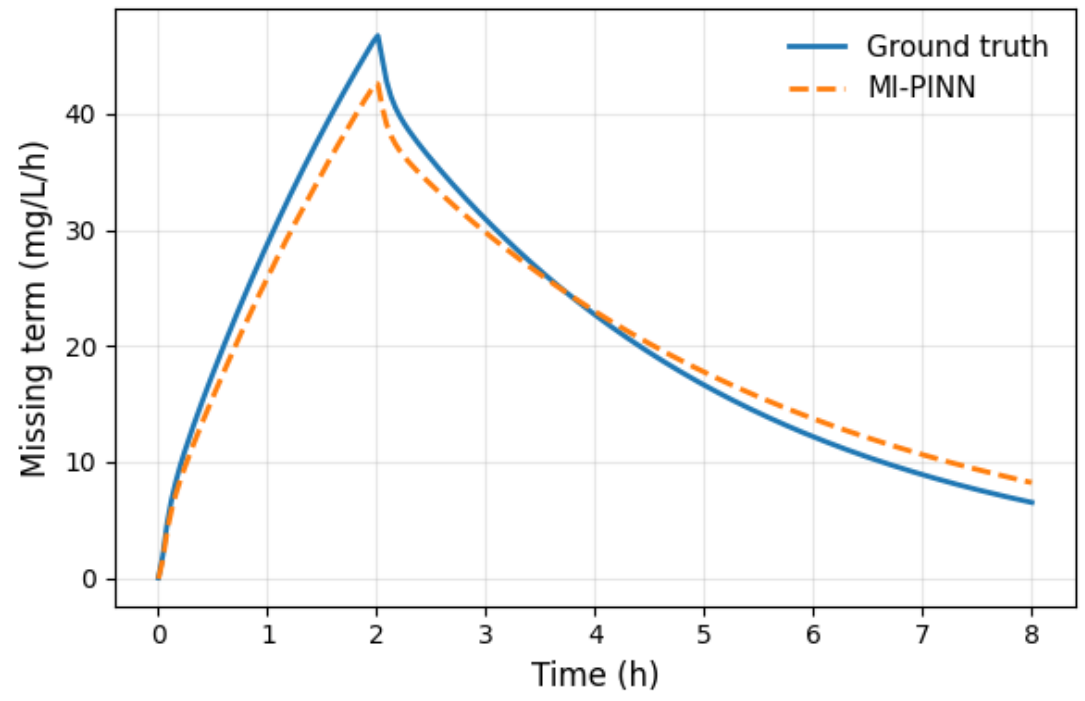

Fig. 7 Prediction of the missing mechanistic term as approximated by the neural network.

With the learned relationship, it is further possible to leverage domain knowledge to guide an algorithmic search towards uncovering the explicit functional form of the missing term. We consider the real-world scenario where the domain knowledge experts are cognizant of the individual terms that make up the masked function but are unaware of how to assemble the terms together. Firstly, we utilize the following grouping of terms based on their physical interpretation

$$X_0 = V_{\max,P} \times C_{\text{Liver}} \tag{19}$$

$$X_1 = M_W \tag{20}$$

$$X_2 = f_{u,b} \times C_{\text{Liver}} \tag{21}$$

$$X_3 = K_{m,\,P} \times f_{u,\,\mathrm{mic}} \tag{22}$$

$$X_4 = MPPGL \times W_{\mathrm{Liver}} \tag{23}$$

$$X_5 = f_{u,\,b} \times R_b \big/ Kp_{\mathrm{Liver}} \tag{24}$$

$$X_6 = 1/V_{\mathrm{Liver}} \tag{25}$$

Based on domain knowledge in binding kinetics, PBPK modeling, and *IVIVE*, we understand that $V_{\max,\,P}$ and $C_{\mathrm{Liver}}$ should always be multiplied together (Eq. (19)). Similarly, only the unbound portion of the drug is subject to metabolism (Eq. (21)). $f_{u,\,\mathrm{mic}}$ is also a term that is only relevant to $K_{m,\,P}$ (Eq. (22)). Eq. (23) is the classical method for performing *IVIVE* to scale clearance for real systems while Eq. (24) is a well-known equation for modeling drug distribution and tissue/blood/plasma partitioning in PBPK models. Lastly, the ODEs as constructed are represented as amount over time instead of concentration over time, hence the division by $V_{\mathrm{Liver}}$ (Eq. (25)) is required for dimensional consistency.

Subsequently, we can use Buckingham Pi Theorem [47] to reduce the following expression for $Y = f\left(X_0, X_1, X_2, X_3, X_4, X_5, X_6\right)$ into the following expression, $\prod_y = \phi\left(\prod_1, \prod_2\right)$, through dimensional analysis.

$$\prod_y = \frac{Y}{X_0 X_5}\frac{X_3}{X_4 X_6} \tag{26}$$

$$\prod_1 = \frac{X_1 X_3}{X_4 X_6} \tag{27}$$

$$\prod_2 = \frac{X_2}{X_4 X_6} \tag{28}$$

By applying symbolic regression to the outputs of the MI-PINN model, we obtain the following

$$\prod_y = \phi\left(\prod_1, \prod_2\right) = 1 \tag{29}$$

$$Y = \frac{X_0}{X_3} X_4 X_5 X_6 \tag{30}$$

where the model is able to correctly deduce the functional form of the missing term (Eq. (30)), absent $X_2/X_1$ that is present in the denominator of the ground truth expression (Eq. (18)). This is perfectly reasonable since $C_{\mathrm{Liver}} << K_{m,\,P}$ during the entire simulation process in the IV dosing scenario, which means that $K_{m,\,P}$ dominates the denominator and the Michaelis-Menten kinetics is linear and approximately equal to $\frac{V_{\max,\,P} \times C_{\mathrm{Liver}}}{K_{m,\,P} \times f_{u,\,\mathrm{mic}}}$. Hence, the learned auxiliary network did not model the impact of $X_2/X_1$ since its overall impact on the ODE system is minimal and this is carried through into the derived expression in Eq. (30).

The findings from our missing term discovery work illustrate a unique implementation and promise of AI. For example, recent works on the use of AI for ODE system (e.g. PK) modeling are largely focused on building better QSAR models for ADME parameter predictions [48, 49]. While accurate estimation of ADME parameters is essential for modeling, an equally important yet often neglected problem lies in crafting appropriate mathematical equations or functions to represent the entire system dynamics. In practice, model complexity is constrained by the extent of existing mechanistic knowledge, making performance improvements difficult when underlying mechanisms are poorly described. Model development frequently relies on making assumptions or heuristic choices to approximate unknown processes. In this context, the ability of MI-PINN to approximate missing or unknown mechanisms through learnable dynamic terms and downstream integration with dimensional analysis and symbolic regression highlights its potential to address this long-standing limitation not only in PBPK modeling, but in the field of system identification in ODE systems across different scientific domains.

### D. IV Theophylline PBPK Model for Parameter Inference

To further test the robustness and evaluate the predictive performance of the MI-PINN framework for data from different individuals with realistic biological variation, we consider an IV theophylline PBPK model as an additional study. The model consists of the same 22-ODE system but with a different set of parameter values based on the ADME of theophylline (Supplementary Table S3).

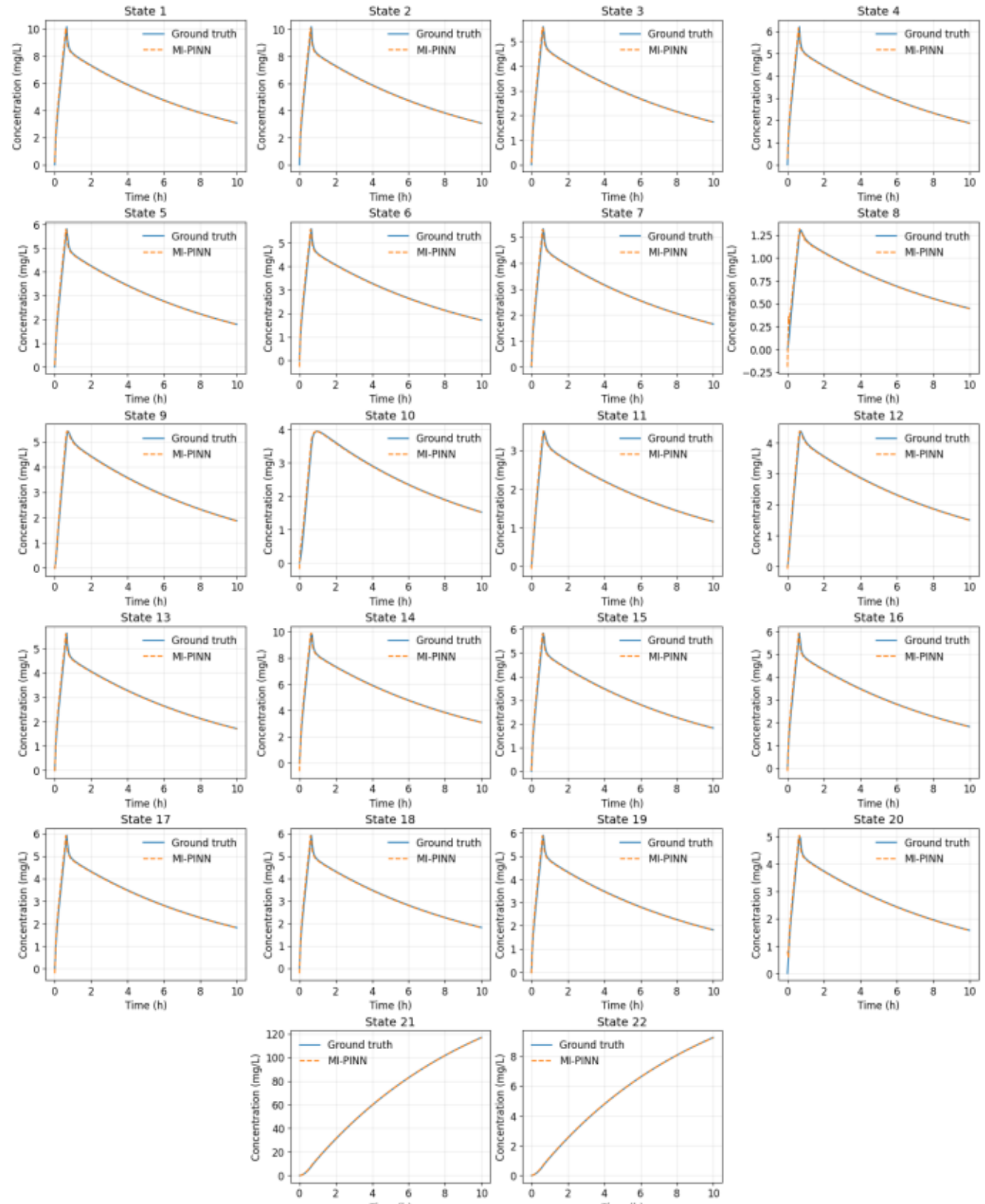

Fig. 8 Prediction results of all 22 state variables obtained via parameter inference for the IV theophylline PBPK model.

Similar to paracetamol, theophylline has relatively straightforward PK and is predominantly metabolized by CYP1A2 and CYP2E1 to form 3-methylxanthine (3-MX), 1-MX and 1, 3-dimethyluric acid (DMU) [50]. Given literature reports that metabolism by CYP1A2 to DMU is the major pathway for theophylline metabolism with the highest clearance kinetics amongst the various CYP isoforms [50, 51], we again mask $V_{\max}$ and $K_m$ for theophylline metabolism by CYP1A2 to DMU, denoted as $V_{\max,T}$ and $K_{m,T}$, respectively.

Following the same training paradigm as in Section V.A, MI-PINN is first meta-trained across varying configurations of $V_{\max,T}$ and $K_{m,T}$. After Stage I, the learned representation $\boldsymbol{\theta}^*$ is kept fixed, and target kinetic parameters are inferred from sparse clinical observations in the second stage. In this study, eight individual venous blood concentration-time profiles, each with a different dosing amount of theophylline ranging from 194 mg to 383 mg infused IV over the same 40 min duration [52], are used for parameter inference, representing inter-subject noise/variability in clinical measurements. In particular, it is worth highlighting that the integration of a mechanistic model in MI-PINN allows for immense flexibility in unifying measurements across disparate experiments (i.e. different subjects each with their own dosing regimens). The prediction results for all 22 state variables obtained via parameter inference are presented in Fig. 8, and the physiological interpretations of State 1-State 22 follow the same definitions as those summarized in TABLE III. The corresponding inferred parameters obtained by MI-PINN and the MATLAB trust region reflective algorithm are summarized in TABLE VII.

TABLE VII
INFERRED KINETIC PARAMETERS FOR THE IV THEOPHYLLINE PBPK MODEL

| Model | $K_{m,T}$ (μmol/L) | $V_{\max,T}$ (μmol/(h·pmol)) | Ratio of $V_{\max,T}$ to $K_{m,T}$ |
|---|---|---|---|
| Ground truth [52] | 230 | $4.39\times10^{-4}$ | $1.91\times10^{-6}$ |
| Trust region reflective algorithm | 999 | $1.92\times10^{-3}$ | $1.92\times10^{-6}$ |
| **MI-PINN** | **214** | **$4.10\times10^{-4}$** | **$1.92\times10^{-6}$** |

As shown in Fig. 8, MI-PINN achieves an MSE of approximately 0.17 across all 22 state variables, indicating accurate prediction performance for the previously unseen theophylline PBPK system. Moreover, similar to the previous two parameter inference results for oral and IV paracetamol, MI-PINN is able to achieve a much more accurate prediction of $V_{\max,T}$ and $K_{m,T}$ compared to the trust region reflective algorithm employed by MATLAB as reported in TABLE VII, despite both having similar $V_{\max,T}/K_{m,T}$ ratio as the ground truth. Consolidating the results seen from TABLE IV, TABLE VI, and TABLE VII, we observe that our MI-PINN consistently provides more accurate prediction of both $V_{\max,T}$ and $K_{m,T}$ compared to the trust region reflective algorithm. While both algorithms are able to fit the respective observed clinical datasets as both are able to provide accurate predictions for the ratio (catalytic efficiency), accurate predictions for the individual $V_{\max,T}$ and $K_{m,T}$ values are also important mechanistically. Under high dosing conditions and/or when combined with high permeability drugs, hepatic CYP enzymes may become saturated and metabolism becomes nonlinear once drug concentration exceeds $K_{m,T}$. Under such circumstances, inaccurate predictions for $K_{m,T}$ may significantly affect the final predicted metabolism kinetics. Likewise, similar issues can arise when predicting transporters' kinetics. Under these conditions, MI-PINN is significantly more valuable for parameter inference. Notably, MI-PINN more accurately infers both $V_{\max,T}$ and $K_{m,T}$ values across all three parameter inference scenarios despite an infinite number of possible combinations of $V_{\max,T}$ and $K_{m,T}$ values that can provide the same catalytic efficiency due to its better learned representation.

## E. Ablation Analysis

(1) Inference Time

For conventional inverse design approaches, numerical ODE solvers are typically invoked repeatedly during gradient-based optimization, which leads to considerable computational overhead. To quantify the efficiency of MI-PINN, we use the 22-ODE IV theophylline PBPK model as an example and compare the wall-clock time of a single forward run using the SciPy odeint solver with that of MI-PINN, as reported in TABLE VIII.

TABLE VIII
WALL-CLOCK INFERENCE TIME PER RUN ON THE THEOPHYLLINE PBPK MODEL

| Model | SciPy | **MI-PINN** |
|---|---|---|
| Inference time/s | 14.66 | **4.17** |

As shown in TABLE VIII, MI-PINN requires only 4.17s for one prediction, compared with 14.66s using SciPy odeint solver, corresponding to a 3.5× speedup. These results demonstrate the computational efficiency of MI-PINN for large-scale inverse PBPK modeling.

(2) Representation Learning Performance

To evaluate the effectiveness of the proposed multi-branch representation learning scheme, we compare it with a single-branch baseline that does not perform clustering based on dynamical similarity. For each setting, we conduct three independent runs with different random initializations. Experiments are carried out on the 22-ODE IV theophylline PBPK model under the forward design setting, and the prediction performance is summarized in TABLE IX.

TABLE IX
PREDICTION RESULTS FOR DIFFERENT LEARNING STRATEGIES ON THE THEOPHYLLINE PBPK MODEL

| Learning strategy | Single-branch learning | **Multi-branch learning** |
|---|---|---|
| MSE | 5.63±7.18 | **0.28±0.25** |

It can be seen from TABLE IX that the proposed adaptive clustering-based multi-branch representation reduces the MSE by approximately 95%, corresponding to about a 20× improvement compared with the single-branch model. This highlights that grouping states with similar dynamics and learning them with dedicated branches can improve representation capacity and modeling accuracy in large heterogeneous PBPK systems. In contrast, a single shared network tends to poorly capture the multi-scale behaviors across different compartments, leading to degraded performance. Overall, these results validate the advantage of the proposed multi-branch representation for PBPK modeling.

(3) Inverse Modeling Performance

To assess the inverse modeling capability of the proposed method, we compare MI-PINN with a conventional PINN baseline. In the conventional PINN, the network weights and the masked physiological parameters are jointly optimized via gradient descent, i.e., parameter inference is coupled with state approximation during training. We consider the 22-ODE IV theophylline PBPK model and infer the key metabolic parameters $K_{m,T}$ and $V_{\max,T}$ in the inverse design setting. For a fair comparison, the baseline PINN adopts the same network architecture and training configuration as MI-PINN described in Section V.D. The quantitative results are summarized in TABLE X.

TABLE X
INVERSE MODELING RESULTS FOR THE IV THEOPHYLLINE PBPK MODEL

| Model | $K_{m,T}$ (μmol/L) | $V_{\max,T}$ (μmol/(h·pmol)) | Ratio of $V_{\max,T}$ to $K_{m,T}$ | MSE |
|---|---|---|---|---|
| Ground truth [52] | 230 | $4.39\times10^{-4}$ | $1.91\times10^{-6}$ | - |
| Conventional PINN | 866 | $1.12\times10^{-4}$ | $1.29\times10^{-7}$ | $1.92\times10^{3}$ |
| **MI-PINN** | **214** | $\mathbf{4.10\times10^{-4}}$ | $\mathbf{1.92\times10^{-6}}$ | **0.17** |

As shown in TABLE X, the conventional PINN exhibits a large estimation error in both state prediction and parameter identification, leading to a high MSE and highly biased inferred parameters. This suggests that directly coupling the optimization of network weights and PBPK parameters can result in a poorly conditioned learning problem in high-dimensional PBPK systems, where multi-scale dynamics and limited observability make the inverse task particularly challenging. In contrast, MI-PINN achieves accurate prediction of all 22 state trajectories with a substantially lower MSE and recovers $K_{m,T}$ and $V_{\max,T}$ values that are close to the ground truth in literature. These results demonstrate that the proposed MI-PINN provides a more reliable and scalable solution for inverse PBPK modeling.

## VI. CONCLUSION

In this work, we propose MI-PINN, a scalable meta-inverse physics-informed learning framework designed to tackle inverse problems in high-dimensional ODE systems under partial observability and sparse data. Using large-scale PBPK modeling as a demanding exemplar, we demonstrate the framework's capability for both parameter inference and missing term discovery. The developed two-stage learning strategy first learns a physics-aware representation from multiple PBPK tasks, and then performs task-level inverse identification under sparse clinical observations. To address the multi-scale dynamics across ODE state variables, we introduce an adaptive clustering-based multi-branch representation that clusters compartments with similar dynamics and enhances the model's representational capacity for multi-scale dynamics. In addition, a physics-informed pseudo-inverse formulation is incorporated to compute the final convex layer in closed form, enabling more reliable inference under noisy observations. Experiments on whole-body paracetamol and theophylline PBPK systems under both IV and oral dosing (22 ODEs and 33 ODEs) demonstrate that MI-PINN can accurately recover masked kinetic parameters and reconstruct missing governing terms, while producing more reliable parameter estimates than solver-based optimization and standard inverse PINN baselines in challenging high-dimensional inverse settings. In future work, MI-PINN can be extended with uncertainty quantification to provide calibrated confidence intervals for inferred parameters and discovered missing terms.

# Supplementary Material

## Table S1. PBPK model parameters for paracetamol

| Parameters | Value | Reference |
|---|---|---|
| **Physiochemical properties** | | |
| Molecular weight (g/mol) | 151.7 | PubChem |
| $f_{u,b}$ | 0.820 | Ladumor *et al.* 2019[1] |
| $R_b$ | 1.580 | |
| LogP | 0.510 | |
| $pK_a$ | 9.460 | |
| Compound type | Monoprotic acid | - |
| **Absorption** | | |
| $P_{eff,man}$ (m/h) | 0.0432 | Ladumor *et al.* 2019[1] |
| Intrinsic solubility (mg/L) | 13650 | |
| **Distribution** | | |
| $V_d$ (L/kg) | 0.900 | Ladumor *et al.* 2019[1] |
| $Kp_{Adipose}$ | 0.375 | Calculated from $V_d$ |
| $Kp_{Bone}$ | 0.898 | |
| $Kp_{Brain}$ | 1.334 | |
| $Kp_{Gut}$ | 1.238 | |
| $Kp_{Heart}$ | 1.110 | |
| $Kp_{Kidney}$ | 1.114 | |
| $Kp_{Liver}$ | 1.151 | |
| $Kp_{Lung}$ | 1.081 | |
| $Kp_{Muscle}$ | 1.022 | |
| $Kp_{Pancreas}$ | 1.199 | |
| $Kp_{RestOfBody}$ | 1.022 | |
| $Kp_{Skin}$ | 1.069 | |
| $Kp_{Spleen}$ | 1.148 | |
| **Metabolism** | | |
| $f_{u,\text{incubation}}$ | 0.922 | Predicted |
| UGT1A1 $V_{max}$ (μmol/h/mg protein) | 0.323 | Ladumor *et al.* 2019[1] |
| UGT1A1 $K_m$ (μM) | 5500 | |
| UGT1A9 $V_{max}$ (μmol/h/mg protein) | 0.540 | |
| UGT1A9 $K_m$ (μM) | 9200 | |
| UGT2B15 $V_{max}$ (μmol/h/mg protein) | 1.801 | |
| UGT2B15 $K_m$ (μM) | 23000 | |
| SULT1A1 $V_{max}$ (μmol/h/mg protein) | 0.0769 | |
| SULT1A1 $K_m$ (μM) | 2400 | |
| SULT1A3 $V_{max}$ (μmol/h/mg protein) | 0.0115 | |
| SULT1A3 $K_m$ (μM) | 1500 | |
| SULT1E1 $V_{max}$ (μmol/h/mg protein) | 0.00830 | |
| SULT1E1 $K_m$ (μM) | 1900 | |
| SULT2A1 $V_{max}$ (μmol/h/mg protein) | 0.0404 | |
| SULT2A1 $K_m$ (μM) | 3700 | |
| CYP1A2 $V_{max}$ (μmol/h/mg protein) | 0.00168 | |
| CYP1A2 $K_m$ (μM) | 220 | |
| CYP2C9 $V_{max}$ (μmol/h/mg protein) | 0.000479 | |
| CYP2C9 $K_m$ (μM) | 660 | |
| CYP2C19 $V_{max}$ (μmol/h/mg protein) | 0.00145 | |
| CYP2C19 $K_m$ (μM) | 2000 | |
| CYP2D6 $V_{max}$ (μmol/h/mg protein) | 0.000319 | |
| CYP2D6 $K_m$ (μM) | 440 | |
| CYP2E1 $V_{max}$ (μmol/h/mg protein) | 0.00437 | |
| CYP2E1 $K_m$ (μM) | 4020 | |
| CYP3A4 $V_{max}$ (μmol/h/mg protein) | 0.00302 | |
| CYP3A4 $K_m$ (μM) | 130 | |
| **Excretion** | | |
| $CL_R$ (L/h) | 1.379 | Ladumor *et al.* 2019[1] |

## Table S2. PBPK model parameters for theophylline

| Parameters | Value | Reference |
|---|---|---|
| **Physiochemical properties** | | |
| Molecular weight (g/mol) | 180.2 | PubChem |
| $f_{u,b}$ | 0.500 | Lombardo *et al.* 2004[2] |
| $R_b$ | 0.815 | Habib *et al.* 1987[3] |
| LogP | -0.020 | Hansch *et al.* 1995[4] |
| $pK_a$ 1 | 8.800 | PubChem |
| $pK_a$ 2 | 0.990 | PubChem |
| Compound type | Zwitterion | - |
| **Absorption** | | |
| $P_{eff,man}$ (m/h) | 0.0178 | Kus *et al.* 2023[5] |
| Intrinsic solubility (mg/L) | 6660 | Li *et al.* 2024[6] |
| **Distribution** | | |
| $V_d$ (L/kg) | 0.500 | Peck *et al.* 1985[7] |
| $Kp_{Adipose}$ | 0.117 | Calculated from $V_d$ |
| $Kp_{Bone}$ | 0.303 | |
| $Kp_{Brain}$ | 0.458 | |
| $Kp_{Gut}$ | 0.482 | |
| $Kp_{Heart}$ | 0.474 | |
| $Kp_{Kidney}$ | 0.454 | |
| $Kp_{Liver}$ | 0.431 | |
| $Kp_{Lung}$ | 0.497 | |
| $Kp_{Muscle}$ | 0.393 | |
| $Kp_{Pancreas}$ | 0.437 | |
| $Kp_{RestOfBody}$ | 0.393 | |
| $Kp_{Skin}$ | 0.489 | |
| $Kp_{Spleen}$ | 0.453 | |
| **Metabolism** | | |
| $f_{u,incubation}$ | 0.931 | Predicted |
| CYP1A2 (→1-MX) $V_{max}$ (μmol/h/pmol) | $3.60 \times 10^{-4}$ | Ha *et al.* 1995[8] |
| CYP1A2 (→1-MX) $K_m$ (μM) | 380 | |
| CYP1A2 (→3-MX) $V_{max}$ (μmol/h/pmol) | $1.48 \times 10^{-4}$ | |
| CYP1A2 (→3-MX) $K_m$ (μM) | 1090 | |
| CYP1A2 (→DMU) $V_{max}$ (μmol/h/pmol) | $4.39 \times 10^{-4}$ | |
| CYP1A2 (→DMU) $K_m$ (μM) | 230 | |
| CYP2E1 (→DMU) $V_{max}$ (μmol/h/pmol) | $4.12 \times 10^{-3}$ | |
| CYP2E1 (→DMU) $K_m$ (μM) | 15300 | |
| ISEF CYP (→DMU) | 0.500 | Optimized |
| **Excretion** | | |
| $CL_R$ (L/h) | 0.310 | Abduljalil *et al.* 2022[9] |

## Table S3. ODEs list for 22 ODE-system (IV scenario)

| Description | ODEs |
|---|---|
| **General ODEs** | |
| Distribution in venous blood | $d(C_{VenousBlood})/dt = 1/V_{VenousBlood}*(-(Q_{Lung}*C_{VenousBlood}) + ((Q_{Brain}*C_{Brain}*Rb)/(Kp_{Brain})) + ((Q_{Heart}*C_{Heart}*Rb)/(Kp_{Heart})) + ((Q_{Liver}*C_{Liver}*Rb)/(Kp_{Liver})) + ((Q_{Adipose}*C_{Adipose}*Rb)/(Kp_{Adipose})) + ((Q_{Skin}*C_{Skin}*Rb)/(Kp_{Skin})) + ((Q_{Muscle}*C_{Muscle}*Rb)/(Kp_{Muscle})) + ((Q_{Bone}*C_{Bone}*Rb)/(Kp_{Bone})) + ((Q_{RestOfBody}*C_{RestOfBody}*Rb)/(Kp_{RestOfBody})) + ((Q_{Kidney}*C_{Kidney}*Rb)/(Kp_{Kidney})))$ |
| Distribution in arterial blood | $d(C_{ArterialBlood})/dt = 1/V_{ArterialBlood}*(((Q_{Lung}*C_{Lung}*R_b)/(Kp_{Lung})) - (Q_{Brain}*C_{ArterialBlood}) - (Q_{Heart}*C_{ArterialBlood}) - (Q_{Stomach}*C_{ArterialBlood}) - (Q_{Spleen}*C_{ArterialBlood}) - (Q_{Pancreas}*C_{ArterialBlood}) - (Q_{HepaticArtery}*C_{ArterialBlood}) - (Q_{Adipose}*C_{ArterialBlood}) - (Q_{Skin}*C_{ArterialBlood}) - (Q_{Muscle}*C_{ArterialBlood}) - (Q_{Bone}*C_{ArterialBlood}) - (Q_{RestOfBody}*C_{ArterialBlood}) - (Q_{Kidney}*C_{ArterialBlood}) - (Q_{Duodenum}*C_{ArterialBlood}) - (Q_{Jejunum}*C_{ArterialBlood}) - (Q_{Ileum}*C_{ArterialBlood}) - (Q_{LargeIntestine}*C_{ArterialBlood}))$ |
| Distribution in brain | $d(C_{Brain})/dt = 1/V_{Brain}*((Q_{Brain}*C_{ArterialBlood}) - ((Q_{Brain}*C_{Brain}*R_b)/(Kp_{Brain})))$ |
| Distribution in lung | $d(C_{Lung})/dt = 1/V_{Lung}*((Q_{Lung}*C_{ArterialBlood}) - ((Q_{Lung}*C_{Lung}*R_b)/(Kp_{Lung})))$ |
| Distribution in heart | $d(C_{Heart})/dt = 1/V_{Heart}*((Q_{Heart}*C_{ArterialBlood}) - ((Q_{Heart}*C_{Heart}*R_b)/(Kp_{Heart})))$ |
| Distribution in spleen | $d(C_{Spleen})/dt = 1/V_{Spleen}*((Q_{Spleen}*C_{ArterialBlood}) - ((Q_{Spleen}*C_{Spleen}*R_b)/(Kp_{Spleen})))$ |
| Distribution in pancreas | $d(C_{Pancreas})/dt = 1/V_{Pancreas}*((Q_{Pancreas}*C_{ArterialBlood}) - ((Q_{Pancreas}*C_{Pancreas}*R_b)/(Kp_{Pancreas})))$ |
| Distribution in adipose | $d(C_{Adipose})/dt = 1/V_{Adipose}*((Q_{Adipose}*C_{ArterialBlood}) - ((Q_{Adipose}*C_{Adipose}*R_b)/(Kp_{Adipose})))$ |
| Distribution in skin | $d(C_{Skin})/dt = 1/V_{Skin}*((Q_{Skin}*C_{ArterialBlood}) - ((Q_{Skin}*C_{Skin}*R_b)/(Kp_{Skin})))$ |
| Distribution in muscle | $d(C_{Muscle})/dt = 1/V_{Muscle}*((Q_{Muscle}*C_{ArterialBlood}) - ((Q_{Muscle}*C_{Muscle}*R_b)/(Kp_{Muscle})))$ |
| Distribution in bone | $d(C_{Bone})/dt = 1/V_{Bone}*((Q_{Bone}*C_{ArterialBlood}) - ((Q_{Bone}*C_{Bone}*R_b)/(Kp_{Bone})))$ |
| Distribution in rest of body | $d(C_{RestOfBody})/dt = 1/V_{RestOfBody}*((Q_{RestOfBody}*C_{ArterialBlood}) - ((Q_{RestOfBody}*C_{RestOfBody}*R_b)/(Kp_{RestOfBody})))$ |
| Distribution + renal clearance in kidney | $d(C_{Kidney})/dt = 1/V_{Kidney}*((Q_{Kidney}*C_{ArterialBlood}) - ((Q_{Kidney}*C_{Kidney}*R_b)/(Kp_{Kidney})) - (CL_R*C_{Kidney}))$ |
| Renally excreted drug in urine | $d(C_{Urine})/dt = 1/V_{Urine}*((CL_R*C_{Kidney}))$ |
| Distribution in portal vein | $d(C_{PortalVein})/dt = 1/V_{PortalVein}*(((Q_{Stomach}*C_{Stomach}*R_b)/(Kp_{Gut})) + ((Q_{Spleen}*C_{Spleen}*R_b)/(Kp_{Spleen})) + ((Q_{Pancreas}*C_{Pancreas}*R_b)/(Kp_{Pancreas})) - (Q_{PortalVein}*C_{PortalVein}) + ((Q_{Duodenum}*C_{DuodenumWall}*R_b)/(Kp_{Gut})) + ((Q_{Jejunum}*C_{JejunumWall}*R_b)/(Kp_{Gut})) + ((Q_{Ileum}*C_{IleumWall}*R_b)/(Kp_{Gut})) + ((Q_{LargeIntestine}*C_{LargeIntestineWall}*R_b)/(Kp_{Gut})))$ |
| Distribution in stomach | $d(C_{Stomach})/dt = 1/V_{Stomach}*((Q_{Stomach}*C_{ArterialBlood}) - ((Q_{Stomach}*C_{Stomach}*R_b)/(Kp_{Gut})))$ |
| Distribution in duodenum | $d(C_{DuodenumWall})/dt = 1/V_{DuodenumWall}*((Q_{Duodenum}*C_{ArterialBlood}) - ((Q_{Duodenum}*C_{DuodenumWall}*R_b)/(Kp_{Gut})))$ |
| Distribution in jejunum | $d(C_{JejunumWall})/dt = 1/V_{JejunumWall}*((Q_{Jejunum}*C_{ArterialBlood}) - ((Q_{Jejunum}*C_{JejunumWall}*R_b)/(Kp_{Gut})))$ |
| Distribution in ileum | $d(C_{IleumWall})/dt = 1/V_{IleumWall}*((Q_{Ileum}*C_{ArterialBlood}) - ((Q_{Ileum}*C_{IleumWall}*R_b)/(Kp_{Gut})))$ |
| Distribution in large intestine | $d(C_{LargeIntestineWall})/dt = 1/V_{LargeIntestineWall}*((Q_{LargeIntestine}*C_{ArterialBlood}) - ((Q_{LargeIntestine}*C_{LargeIntestineWall}*R_b)/(Kp_{Gut})))$ |
| **Drug-specific ODEs** | |
| Paracetamol metabolism + distribution in liver | $d(C_{Liver})/dt = 1/V_{Liver}*((Q_{PortalVein}*C_{PortalVein}) + (Q_{HepaticArtery}*C_{ArterialBlood}) - ((Q_{Liver}*C_{Liver}*R_b)/(Kp_{Liver}))$ - <br> $((V_{max}(UGT1A9)/((K_m(UGT1A9)*f_{u,mic})+((C_{Liver}/M_w)*f_{u,b}))*MPPGL*W_{Liver})*(C_{Liver}*f_{u,b})*R_b/Kp_{Liver})$ - <br> $((V_{max}(UGT1A1)/((K_m(UGT1A1)*f_{u,mic})+((C_{Liver}/M_w)*f_{u,b}))*MPPGL*W_{Liver})*(C_{Liver}*f_{u,b})*R_b/Kp_{Liver})$ - <br> $((V_{max}(UGT2B15)/((K_m(UGT2B15)*f_{u,mic})+((C_{Liver}/M_w)*f_{u,b}))*MPPGL*W_{Liver})*(C_{Liver}*f_{u,b})*R_b/Kp_{Liver})$ - <br> $((V_{max}(SULT1A1)/((K_m(SULT1A1)*f_{u,mic})+((C_{Liver}/M_w)*f_{u,b}))*CPPGL*W_{Liver})*(C_{Liver}*f_{u,b})*R_b/Kp_{Liver})$ - <br> $((V_{max}(SULT1A3)/((K_m(SULT1A3)*f_{u,mic})+((C_{Liver}/M_w)*f_{u,b}))*CPPGL*W_{Liver})*(C_{Liver}*f_{u,b})*R_b/Kp_{Liver})$ - <br> $((V_{max}(SULT1E1)/((K_m(SULT1E1)*f_{u,mic})+((C_{Liver}/M_w)*f_{u,b}))*CPPGL*W_{Liver})*(C_{Liver}*f_{u,b})*R_b/Kp_{Liver})$ - <br> $((V_{max}(SULT2A1)/((K_m(SULT2A1)*f_{u,mic})+((C_{Liver}/M_w)*f_{u,b}))*CPPGL*W_{Liver})*(C_{Liver}*f_{u,b})*R_b/Kp_{Liver})$ - <br> $((V_{max}(CYP1A2)/((K_m(CYP1A2)*f_{u,mic})+((C_{Liver}/M_w)*f_{u,b}))*MPPGL*W_{Liver})*(C_{Liver}*f_{u,b})*R_b/Kp_{Liver})$ - <br> $((V_{max}(CYP2C9)/((K_m(CYP2C9)*f_{u,mic})+((C_{Liver}/M_w)*f_{u,b}))*MPPGL*W_{Liver})*(C_{Liver}*f_{u,b})*R_b/Kp_{Liver})$ - <br> $((V_{max}(CYP2C19)/((K_m(CYP2C19)*f_{u,mic})+((C_{Liver}/M_w)*f_{u,b}))*MPPGL*W_{Liver})*(C_{Liver}*f_{u,b})*R_b/Kp_{Liver})$ - <br> $((V_{max}(CYP2D6)/((K_m(CYP2D6)*f_{u,mic})+((C_{Liver}/M_w)*f_{u,b}))*MPPGL*W_{Liver})*(C_{Liver}*f_{u,b})*R_b/Kp_{Liver})$ - <br> $((V_{max}(CYP2E1)/((K_m(CYP2E1)*f_{u,mic})+((C_{Liver}/M_w)*f_{u,b}))*MPPGL*W_{Liver})*(C_{Liver}*f_{u,b})*R_b/Kp_{Liver})$ - <br> $((V_{max}(CYP3A4)/((K_m(CYP3A4)*f_{u,mic})+((C_{Liver}/M_w)*f_{u,b}))*MPPGL*W_{Liver})*(C_{Liver}*f_{u,b})*R_b/Kp_{Liver}))$ |
| Paracetamol metabolite formation in liver | $d(C_{Liver}(Metabolite))/dt = 1/V_{Liver}*(((V_{max}(UGT1A9)/((K_m(UGT1A9)*f_{u,mic})+((C_{Liver}/M_w)*f_{u,b}))*MPPGL*W_{Liver})*(C_{Liver}*f_{u,b})*R_b/Kp_{Liver})$ + <br> $((V_{max}(UGT1A1)/((K_m(UGT1A1)*f_{u,mic})+((C_{Liver}/M_w)*f_{u,b}))*MPPGL*W_{Liver})*(C_{Liver}*f_{u,b})*R_b/Kp_{Liver})$ + <br> $((V_{max}(UGT2B15)/((K_m(UGT2B15)*f_{u,mic})+((C_{Liver}/M_w)*f_{u,b}))*MPPGL*W_{Liver})*(C_{Liver}*f_{u,b})*R_b/Kp_{Liver})$ + <br> $((V_{max}(SULT1A1)/((K_m(SULT1A1)*f_{u,mic})+((C_{Liver}/M_w)*f_{u,b}))*CPPGL*W_{Liver})*(C_{Liver}*f_{u,b})*R_b/Kp_{Liver})$ + <br> $((V_{max}(SULT1A3)/((K_m(SULT1A3)*f_{u,mic})+((C_{Liver}/M_w)*f_{u,b}))*CPPGL*W_{Liver})*(C_{Liver}*f_{u,b})*R_b/Kp_{Liver})$ + <br> $((V_{max}(SULT1E1)/((K_m(SULT1E1)*f_{u,mic})+((C_{Liver}/M_w)*f_{u,b}))*CPPGL*W_{Liver})*(C_{Liver}*f_{u,b})*R_b/Kp_{Liver})$ + <br> $((V_{max}(SULT2A1)/((K_m(SULT2A1)*f_{u,mic})+((C_{Liver}/M_w)*f_{u,b}))*CPPGL*W_{Liver})*(C_{Liver}*f_{u,b})*R_b/Kp_{Liver})$ + <br> $((V_{max}(CYP1A2)/((K_m(CYP1A2)*f_{u,mic})+((C_{Liver}/M_w)*f_{u,b}))*MPPGL*W_{Liver})*(C_{Liver}*f_{u,b})*R_b/Kp_{Liver})$ + <br> $((V_{max}(CYP2C9)/((K_m(CYP2C9)*f_{u,mic})+((C_{Liver}/M_w)*f_{u,b}))*MPPGL*W_{Liver})*(C_{Liver}*f_{u,b})*R_b/Kp_{Liver})$ + <br> $((V_{max}(CYP2C19)/((K_m(CYP2C19)*f_{u,mic})+((C_{Liver}/M_w)*f_{u,b}))*MPPGL*W_{Liver})*(C_{Liver}*f_{u,b})*R_b/Kp_{Liver})$ + <br> $((V_{max}(CYP2D6)/((K_m(CYP2D6)*f_{u,mic})+((C_{Liver}/M_w)*f_{u,b}))*MPPGL*W_{Liver})*(C_{Liver}*f_{u,b})*R_b/Kp_{Liver})$ + <br> $((V_{max}(CYP2E1)/((K_m(CYP2E1)*f_{u,mic})+((C_{Liver}/M_w)*f_{u,b}))*MPPGL*W_{Liver})*(C_{Liver}*f_{u,b})*R_b/Kp_{Liver})$ + <br> $((V_{max}(CYP3A4)/((K_m(CYP3A4)*f_{u,mic})+((C_{Liver}/M_w)*f_{u,b}))*MPPGL*W_{Liver})*(C_{Liver}*f_{u,b})*R_b/Kp_{Liver}))$ |
| Theophylline metabolism + distribution in liver | $d(C_{Liver})/dt = 1/V_{Liver}*((Q_{PortalVein}*C_{PortalVein}) + (Q_{HepaticArtery}*C_{ArterialBlood}) - ((Q_{Liver}*C_{Liver}*R_b)/(Kp_{Liver}))$ - <br> $((CYP1A2(1MX)_{ISEF}*V_{max}(CYP1A2(1MX))*CYP1A2_{Abundance,Liver})/(K_m(CYP1A2(1MX))*f_{u,mic}+((C_{Liver}/M_w)*f_{u,b}))*MPPGL*W_{Liver}*(C_{Liver}*f_{u,b})*R_b/Kp_{Liver}$ <br> $)$ - <br> $((CYP1A2(3MX)_{ISEF}*V_{max}(CYP1A2(3MX))*CYP1A2_{Abundance,Liver})/(K_m(CYP1A2(3MX))*f_{u,mic}+((C_{Liver}/M_w)*f_{u,b}))*MPPGL*W_{Liver}*(C_{Liver}*f_{u,b})*R_b/Kp_{Liver}$ |
|  | $) - ((DMU_{ISEF}*V_{max}(CYP1A2(DMU))*CYP1A2_{Abundance,Liver})/(K_m(CYP1A2(DMU))*f_{u,mic}+((C_{Liver}/M_w)*f_{u,b}))*MPPGL*W_{Liver}*(C_{Liver}*f_{u,b})*R_b/Kp_{Liver}) - ((DMU_{ISEF}*V_{max}(CYP2E1(DMU))*CYP2E1_{Abundance,Liver})/(K_m(CYP2E1(DMU))*f_{u,mic}+((C_{Liver}/M_w)*f_{u,b}))*MPPGL*W_{Liver}*(C_{Liver}*f_{u,b})*R_b/Kp_{Liver}))$ |
| Theophylline metabolite formation in liver | $d(C_{Liver}(Metabolite))/dt = 1/V_{Liver}*(((CYP1A2(1MX)_{ISEF}*V_{max}(CYP1A2(1MX))*CYP1A2_{Abundance,Liver})/(K_m(CYP1A2(1MX))*f_{u,mic}+((C_{Liver}/M_w)*f_{u,b}))*MPPGL*W_{Liver}*(C_{Liver}*f_{u,b})*R_b/Kp_{Liver}) + ((CYP1A2(3MX)_{ISEF}*V_{max}(CYP1A2(3MX))*CYP1A2_{Abundance,Liver})/(K_m(CYP1A2(3MX))*f_{u,mic}C_{Liver}/M_w)*f_{u,b}))*MPPGL*W_{Liver}*(C_{Liver}*f_{u,b})*R_b/Kp_{Liver}) + ((DMU_{ISEF}*V_{max}(CYP1A2(DMU))*CYP1A2_{Abundance,Liver})/(K_m(CYP1A2(DMU))*f_{u,mic}+((C_{Liver}/M_w)*f_{u,b}))*MPPGL*W_{Liver}*(C_{Liver}*f_{u,b})*R_b/Kp_{Liver}) + ((DMU_{ISEF}*V_{max}(CYP2E1(DMU))*CYP2E1_{Abundance,Liver})/(K_m(CYP2E1(DMU))*f_{u,mic}+((C_{Liver}/M_w)*f_{u,b}))*MPPGL*W_{Liver}*(C_{Liver}*f_{u,b})*R_b/Kp_{Liver}))$ |

## Table S4. ODEs list for 33 ODE-system (oral scenario)

| Description | ODEs |
|---|---|
| **General ODEs** | |
| Distribution in venous blood | $d(C_{VenousBlood})/dt = 1/V_{VenousBlood}*(-(Q_{Lung}*C_{VenousBlood}) + ((Q_{Brain}*C_{Brain}*R_b)/(Kp_{Brain})) + ((Q_{Heart}*C_{Heart}*R_b)/(Kp_{Heart})) + ((Q_{Liver}*C_{Liver}*R_b)/(Kp_{Liver})) + ((Q_{Adipose}*C_{Adipose}*R_b)/(Kp_{Adipose})) + ((Q_{Skin}*C_{Skin}*R_b)/(Kp_{Skin})) + ((Q_{Muscle}*C_{Muscle}*R_b)/(Kp_{Muscle})) + ((Q_{Bone}*C_{Bone}*R_b)/(Kp_{Bone})) + ((Q_{RestOfBody}*C_{RestOfBody}*R_b)/(Kp_{RestOfBody})) + ((Q_{Kidney}*C_{Kidney}*R_b)/(Kp_{Kidney})))$ |
| Distribution in arterial blood | $d(C_{ArterialBlood})/dt = 1/V_{ArterialBlood}*(((Q_{Lung}*C_{Lung}*R_b)/(Kp_{Lung})) - (Q_{Brain}*C_{ArterialBlood}) - (Q_{Heart}*C_{ArterialBlood}) - (Q_{Stomach}*C_{ArterialBlood}) - (Q_{Spleen}*C_{ArterialBlood}) - (Q_{Pancreas}*C_{ArterialBlood}) - (Q_{HepaticArtery}*C_{ArterialBlood}) - (Q_{Adipose}*C_{ArterialBlood}) - (Q_{Skin}*C_{ArterialBlood}) - (Q_{Muscle}*C_{ArterialBlood}) - (Q_{Bone}*C_{ArterialBlood}) - (Q_{RestOfBody}*C_{ArterialBlood}) - (Q_{Kidney}*C_{ArterialBlood}) - (Q_{Duodenum}*C_{ArterialBlood}) - (Q_{Jejunum}*C_{ArterialBlood}) - (Q_{Ileum}*C_{ArterialBlood}) - (Q_{LargeIntestine}*C_{ArterialBlood}))$ |
| Distribution in brain | $d(C_{Brain})/dt = 1/V_{Brain}*((Q_{Brain}*C_{ArterialBlood}) - ((Q_{Brain}*C_{Brain}*R_b)/(Kp_{Brain})))$ |
| Distribution in lung | $d(C_{Lung})/dt = 1/V_{Lung}*((Q_{Lung}*C_{VenousBlood}) - ((Q_{Lung}*C_{Lung}*R_b)/(Kp_{Lung})))$ |
| Distribution in heart | $d(C_{Heart})/dt = 1/V_{Heart}*((Q_{Heart}*C_{ArterialBlood}) - ((Q_{Heart}*C_{Heart}*R_b)/(Kp_{Heart})))$ |
| Distribution in spleen | $d(C_{Spleen})/dt = 1/V_{Spleen}*((Q_{Spleen}*C_{ArterialBlood}) - ((Q_{Spleen}*C_{Spleen}*R_b)/(Kp_{Spleen})))$ |
| Distribution in pancreas | $d(C_{Pancreas})/dt = 1/V_{Pancreas}*((Q_{Pancreas}*C_{ArterialBlood}) - ((Q_{Pancreas}*C_{Pancreas}*R_b)/(Kp_{Pancreas})))$ |
| Distribution in adipose | $d(C_{Adipose})/dt = 1/V_{Adipose}*((Q_{Adipose}*C_{ArterialBlood}) - ((Q_{Adipose}*C_{Adipose}*R_b)/(Kp_{Adipose})))$ |
| Distribution in skin | $d(C_{Skin})/dt = 1/V_{Skin}*((Q_{Skin}*C_{ArterialBlood}) - ((Q_{Skin}*C_{Skin}*R_b)/(Kp_{Skin})))$ |
| Distribution in muscle | $d(C_{Muscle})/dt = 1/V_{Muscle}*((Q_{Muscle}*C_{ArterialBlood}) - ((Q_{Muscle}*C_{Muscle}*R_b)/(Kp_{Muscle})))$ |
| Distribution in bone | $d(C_{Bone})/dt = 1/V_{Bone}*((Q_{Bone}*C_{ArterialBlood}) - ((Q_{Bone}*C_{Bone}*R_b)/(Kp_{Bone})))$ |
| Distribution in rest of body | $d(C_{RestOfBody})/dt = 1/V_{RestOfBody}*((Q_{RestOfBody}*C_{ArterialBlood}) - ((Q_{RestOfBody}*C_{RestOfBody}*R_b)/(Kp_{RestOfBody})))$ |
| Distribution + renal clearance in kidney | $d(C_{Kidney})/dt = 1/V_{Kidney}*((Q_{Kidney}*C_{ArterialBlood}) - ((Q_{Kidney}*C_{Kidney}*R_b)/(Kp_{Kidney})) - (CL_R*C_{Kidney}))$ |
| Renally excreted drug in urine | $d(C_{Urine})/dt = 1/V_{Urine}*((CL_R*C_{Kidney}))$ |
| Distribution in portal vein | $d(C_{PortalVein})/dt = 1/V_{PortalVein}*(((Q_{Stomach}*C_{Stomach}*R_b)/(Kp_{Gut})) + ((Q_{Spleen}*C_{Spleen}*R_b)/(Kp_{Spleen})) + ((Q_{Pancreas}*C_{Pancreas}*R_b)/(Kp_{Pancreas})) - (Q_{PortalVein}*C_{PortalVein}) + ((Q_{Duodenum}*C_{DuodenumWall}*R_b)/(Kp_{Gut})) + ((Q_{Jejunum}*C_{JejunumWall}*R_b)/(Kp_{Gut})) + ((Q_{Ileum}*C_{IleumWall}*R_b)/(Kp_{Gut})) + ((Q_{LargeIntestine}*C_{LargeIntestineWall}*R_b)/(Kp_{Gut})))$ |
| Distribution in stomach | $d(C_{Stomach})/dt = 1/V_{Stomach}*((Q_{Stomach}*C_{ArterialBlood}) - ((Q_{Stomach}*C_{Stomach}*R_b)/(Kp_{Gut}))) + ((ka_{Stomach}*C_{StomachFluid})*V_{StomachFluid}))$ |
| Distribution in duodenum | $d(C_{DuodenumWall})/dt = 1/V_{DuodenumWall}*(((ka_{Duodenum}*C_{DuodenumLumen})*V_{DuodenumLumen}) + (Q_{Duodenum}*C_{ArterialBlood}) - ((Q_{Duodenum}*C_{DuodenumWall}*R_b)/(Kp_{Gut})))$ |
| Distribution in jejunum | $d(C_{JejunumWall})/dt = 1/V_{JejunumWall}*(((ka_{Jejunum}*C_{JejunumLumen})*V_{JejunumLumen}) + (Q_{Jejunum}*C_{ArterialBlood}) - ((Q_{Jejunum}*C_{JejunumWall}*R_b)/(Kp_{Gut})))$ |
| Distribution in ileum | $d(C_{IleumWall})/dt = 1/V_{IleumWall}*(((ka_{Ileum}*C_{IleumLumen})*V_{IleumLumen}) + (Q_{Ileum}*C_{ArterialBlood}) - ((Q_{Ileum}*C_{IleumWall}*R_b)/(Kp_{Gut})))$ |
| Distribution in large intestine | $d(C_{LargeIntestineWall})/dt = 1/V_{LargeIntestineWall}*(((ka_{LargeIntestine}*C_{LargeIntestineLumen})*V_{LargeIntestineLumen}) + (Q_{LargeIntestine}*C_{ArterialBlood}) - ((Q_{LargeIntestine}*C_{LargeIntestineWall}*R_b)/(Kp_{Gut})))$ |
| Drug dissolution + transit in stomach (solid drug) | $d(C(Solid)_{StomachFluid})/dt = -(Dissolution_Parameter*C(Solid)_{StomachFluid}*(Solubility_{Stomach}-C_{StomachFluid})) - (kt_{Stomach}*C(Solid)_{StomachFluid})$ |
| Drug dissolution + transit in stomach (solubilized drug) | $d(C_{StomachFluid})/dt = 1/V_{StomachFluid}*((Dissolution_Parameter*C(Solid)_{StomachFluid}*(Solubility_{Stomach}-C_{StomachFluid})) - ((kt_{Stomach}*C_{StomachFluid})*V_{StomachFluid}) - ((ka_{Stomach}*C_{StomachFluid})*V_{StomachFluid})$ |
| Drug dissolution + transit in duodenum (solid drug) | $d(C(Solid)_{DuodenumLumen})/dt = -(Dissolution_Parameter*C(Solid)_{DuodenumLumen}*(Solubility_{Duodenum}-C_{DuodenumLumen})) + (kt_{Stomach}*C(Solid)_{StomachFluid}) - (kt_{Duodenum}*C(Solid)_{DuodenumLumen})$ |
| Drug dissolution + transit in duodenum (solubilized drug) | $d(C_{DuodenumLumen})/dt = 1/V_{DuodenumLumen}*((Dissolution_Parameter*C(Solid)_{DuodenumLumen}*(Solubility_{Duodenum}-C_{DuodenumLumen})) + ((kt_{Stomach}*C_{StomachFluid})*V_{StomachFluid}) - ((kt_{Duodenum}*C_{DuodenumLumen})*V_{DuodenumLumen}) - ((ka_{Duodenum}*C_{DuodenumLumen})*V_{DuodenumLumen}))$ |
| Drug dissolution + transit in jejunum (solid drug) | $d(C(Solid)_{JejunumLumen})/dt = -(Dissolution_Parameter*C(Solid)_{JejunumLumen}*(Solubility_{Jejunum}-C_{JejunumLumen})) + (kt_{Duodenum}*C(Solid)_{DuodenumLumen}) - (kt_{Jejunum}*C(Solid)_{JejunumLumen})$ |
| Drug dissolution + transit in jejunum (solubilized drug) | $d(C_{JejunumLumen})/dt = 1/V_{JejunumLumen}*((Dissolution_Parameter*C(Solid)_{JejunumLumen}*(Solubility_{Jejunum}-C_{JejunumLumen})) + ((kt_{Duodenum}*C_{DuodenumLumen})*V_{DuodenumLumen}) - ((kt_{Jejunum}*C_{JejunumLumen})*V_{JejunumLumen}) - ((ka_{Jejunum}*C_{JejunumLumen})*V_{JejunumLumen}))$ |
| Drug dissolution + transit in ileum (solid drug) | $d(C(Solid)_{IleumLumen})/dt = -(Dissolution_Parameter*C(Solid)_{IleumLumen}*(Solubility_{Ileum}-C_{IleumLumen})) + (kt_{Jejunum}*C(Solid)_{JejunumLumen}) - (kt_{Ileum}*C(Solid)_{IleumLumen})$ |
| Drug dissolution + transit in ileum (solubilized drug) | $d(C_{IleumLumen})/dt = 1/V_{IleumLumen}*((Dissolution_Parameter*C(Solid)_{IleumLumen}*(Solubility_{Ileum}-C_{IleumLumen})) + ((kt_{Jejunum}*C_{JejunumLumen})*V_{JejunumLumen}) - ((kt_{Ileum}*C_{IleumLumen})*V_{IleumLumen}) - ((ka_{Ileum}*C_{IleumLumen})*V_{IleumLumen}))$ |
| Drug dissolution + transit in large intestine (solid drug) | $d(C(Solid)_{LargeIntestineLumen})/dt = -(Dissolution_Parameter*C(Solid)_{LargeIntestineLumen}*(Solubility_{LargeIntestine}-C_{LargeIntestineLumen})) + (kt_{Ileum}*C(Solid)_{IleumLumen}) - (kt_{LargeIntestine}*C(Solid)_{LargeIntestineLumen})$ |
| Drug dissolution + transit in large intestine (solubilized drug) | $d(C_{LargeIntestineLumen})/dt = 1/V_{LargeIntestineLumen}*((Dissolution_Parameter*C(Solid)_{LargeIntestineLumen}*(Solubility_{LargeIntestine}-C_{LargeIntestineLumen})) + ((kt_{Ileum}*C_{IleumLumen})*V_{IleumLumen}) - ((kt_{LargeIntestine}*C_{LargeIntestineLumen})*V_{LargeIntestineLumen}) - ((ka_{LargeIntestine}*C_{LargeIntestineLumen})*V_{LargeIntestineLumen}))$ |
| Drug amount excreted in feaces | $d(C_{Feaces})/dt = 1/V_{Feaces}*((kt_{LargeIntestine}*C(Solid)_{LargeIntestineLumen}) + ((kt_{LargeIntestine}*C_{LargeIntestineLumen})*V_{LargeIntestineLumen}))$ |
| **Drug-specific ODEs** | |
| Paracetamol metabolism + distribution in liver | $d(C_{Liver})/dt = 1/V_{Liver}*((Q_{PortalVein}*C_{PortalVein}) + (Q_{HepaticArtery}*C_{ArterialBlood}) - ((Q_{Liver}*C_{Liver}*R_b)/(Kp_{Liver})) -$<br>$((V_{max}(UGT1A9)/((K_m(UGT1A9)*f_{u,mic})+((C_{Liver}/M_w)*f_{u,b}))*MPPGL*W_{Liver})*(C_{Liver}*f_{u,b})*R_b/Kp_{Liver}) -$<br>$((V_{max}(UGT1A1)/((K_m(UGT1A1)*f_{u,mic})+((C_{Liver}/M_w)*f_{u,b}))*MPPGL*W_{Liver})*(C_{Liver}*f_{u,b})*R_b/Kp_{Liver}) -$<br>$((V_{max}(UGT2B15)/((K_m(UGT2B15)*f_{u,mic})+((C_{Liver}/M_w)*f_{u,b}))*MPPGL*W_{Liver})*(C_{Liver}*f_{u,b})*R_b/Kp_{Liver}) -$<br>$((V_{max}(SULT1A1)/((K_m(SULT1A1)*f_{u,mic})+((C_{Liver}/M_w)*f_{u,b}))*CPPGL*W_{Liver})*(C_{Liver}*f_{u,b})*R_b/Kp_{Liver}) -$<br>$((V_{max}(SULT1A3)/((K_m(SULT1A3)*f_{u,mic})+((C_{Liver}/M_w)*f_{u,b}))*CPPGL*W_{Liver})*(C_{Liver}*f_{u,b})*R_b/Kp_{Liver}) -$<br>$((V_{max}(SULT1E1)/((K_m(SULT1E1)*f_{u,mic})+((C_{Liver}/M_w)*f_{u,b}))*CPPGL*W_{Liver})*(C_{Liver}*f_{u,b})*R_b/Kp_{Liver}) -$<br>$((V_{max}(SULT2A1)/((K_m(SULT2A1)*f_{u,mic})+((C_{Liver}/M_w)*f_{u,b}))*CPPGL*W_{Liver})*(C_{Liver}*f_{u,b})*R_b/Kp_{Liver}) -$<br>$((V_{max}(CYP1A2)/((K_m(CYP1A2)*f_{u,mic})+((C_{Liver}/M_w)*f_{u,b}))*MPPGL*W_{Liver})*(C_{Liver}*f_{u,b})*R_b/Kp_{Liver}) -$<br>$((V_{max}(CYP2C9)/((K_m(CYP2C9)*f_{u,mic})+((C_{Liver}/M_w)*f_{u,b}))*MPPGL*W_{Liver})*(C_{Liver}*f_{u,b})*R_b/Kp_{Liver}) -$<br>$((V_{max}(CYP2C19)/((K_m(CYP2C19)*f_{u,mic})+((C_{Liver}/M_w)*f_{u,b}))*MPPGL*W_{Liver})*(C_{Liver}*f_{u,b})*R_b/Kp_{Liver}) -$<br>$((V_{max}(CYP2D6)/((K_m(CYP2D6)*f_{u,mic})+((C_{Liver}/M_w)*f_{u,b}))*MPPGL*W_{Liver})*(C_{Liver}*f_{u,b})*R_b/Kp_{Liver}) -$<br>$((V_{max}(CYP2E1)/((K_m(CYP2E1)*f_{u,mic})+((C_{Liver}/M_w)*f_{u,b}))*MPPGL*W_{Liver})*(C_{Liver}*f_{u,b})*R_b/Kp_{Liver}) -$<br>$((V_{max}(CYP3A4)/((K_m(CYP3A4)*f_{u,mic})+((C_{Liver}/M_w)*f_{u,b}))*MPPGL*W_{Liver})*(C_{Liver}*f_{u,b})*R_b/Kp_{Liver}))$ |
| Paracetamol metabolite formation in liver | $d(C_{Liver}(Metabolite))/dt = 1/V_{Liver}*(((V_{max}(UGT1A9)/((K_m(UGT1A9)*f_{u,mic})+((C_{Liver}/M_w)*f_{u,b}))*MPPGL*W_{Liver})*(C_{Liver}*f_{u,b})*R_b/Kp_{Liver}) +$<br>$((V_{max}(UGT1A1)/((K_m(UGT1A1)*f_{u,mic})+((C_{Liver}/M_w)*f_{u,b}))*MPPGL*W_{Liver})*(C_{Liver}*f_{u,b})*R_b/Kp_{Liver}) +$<br>$((V_{max}(UGT2B15)/((K_m(UGT2B15)*f_{u,mic})+((C_{Liver}/M_w)*f_{u,b}))*MPPGL*W_{Liver})*(C_{Liver}*f_{u,b})*R_b/Kp_{Liver}) +$<br>$((V_{max}(SULT1A1)/((K_m(SULT1A1)*f_{u,mic})+((C_{Liver}/M_w)*f_{u,b}))*CPPGL*W_{Liver})*(C_{Liver}*f_{u,b})*R_b/Kp_{Liver}) +$<br>$((V_{max}(SULT1A3)/((K_m(SULT1A3)*f_{u,mic})+((C_{Liver}/M_w)*f_{u,b}))*CPPGL*W_{Liver})*(C_{Liver}*f_{u,b})*R_b/Kp_{Liver}) +$<br>$((V_{max}(SULT1E1)/((K_m(SULT1E1)*f_{u,mic})+((C_{Liver}/M_w)*f_{u,b}))*CPPGL*W_{Liver})*(C_{Liver}*f_{u,b})*R_b/Kp_{Liver}) +$<br>$((V_{max}(SULT2A1)/((K_m(SULT2A1)*f_{u,mic})+((C_{Liver}/M_w)*f_{u,b}))*CPPGL*W_{Liver})*(C_{Liver}*f_{u,b})*R_b/Kp_{Liver}) +$<br>$((V_{max}(CYP1A2)/((K_m(CYP1A2)*f_{u,mic})+((C_{Liver}/M_w)*f_{u,b}))*MPPGL*W_{Liver})*(C_{Liver}*f_{u,b})*R_b/Kp_{Liver}) +$<br>$((V_{max}(CYP2C9)/((K_m(CYP2C9)*f_{u,mic})+((C_{Liver}/M_w)*f_{u,b}))*MPPGL*W_{Liver})*(C_{Liver}*f_{u,b})*R_b/Kp_{Liver}) +$<br>$((V_{max}(CYP2C19)/((K_m(CYP2C19)*f_{u,mic})+((C_{Liver}/M_w)*f_{u,b}))*MPPGL*W_{Liver})*(C_{Liver}*f_{u,b})*R_b/Kp_{Liver}) +$<br>$((V_{max}(CYP2D6)/((K_m(CYP2D6)*f_{u,mic})+((C_{Liver}/M_w)*f_{u,b}))*MPPGL*W_{Liver})*(C_{Liver}*f_{u,b})*R_b/Kp_{Liver}) +$<br>$((V_{max}(CYP2E1)/((K_m(CYP2E1)*f_{u,mic})+((C_{Liver}/M_w)*f_{u,b}))*MPPGL*W_{Liver})*(C_{Liver}*f_{u,b})*R_b/Kp_{Liver}) +$<br>$((V_{max}(CYP3A4)/((K_m(CYP3A4)*f_{u,mic})+((C_{Liver}/M_w)*f_{u,b}))*MPPGL*W_{Liver})*(C_{Liver}*f_{u,b})*R_b/Kp_{Liver}))$ |

**Table S5. Compartments list for paracetamol and theophylline model**

| Compartment | Volume (L) | |
|---|---|---|
| | Paracetamol | Theophylline |
| Arterial blood | 0.966 | 1.051 |
| Venous blood | 3.100 | 3.374 |
| Brain | 1.337 | 1.323 |
| Lung | 1.096 | 0.651 |
| Heart | 0.782 | 0.355 |
| Spleen | 0.136 | 0.149 |
| Pancreas | 0.134 | 0.139 |
| Adipose | 18.970 | 15.148 |
| Skin | 2.877 | 3.759 |
| Muscle | 26.739 | 32.321 |
| Bone | 7.745 | 4.269 |
| Rest of body | 0.324 | 6.836 |
| Kidney | 0.283 | 0.323 |
| Portal vein | 1.000 | 1.000 |
| Stomach | 0.138 | 0.152 |
| Duodenum wall | 0.051 | 0.057 |
| Jejunum wall | 0.258 | 0.284 |
| Ileum wall | 0.285 | 0.314 |
| Large intestine wall | 0.341 | 0.375 |
| Liver | 1.598 | 1.591 |
| Urine | 1.000 | 1.000 |
| **For 33 ODE-system with oral dosing** | | |
| Stomach fluid | 0.045 | NA |
| Duodenum lumen | 0.009 | |
| Jejunum lumen | 0.046 | |
| Ileum lumen | 0.050 | |
| Large intestine lumen | 0.013 | |
| Feaces | 1.000 | |

**Table S6. Systems parameter list for paracetamol and theophylline model**

| Parameter | Unit | Value | |
|---|---|---|---|
| | | Paracetamol | Theophylline |
| $Q_{Brain}$ | liter/hour | 46.800 | 46.361 |
| $Q_{Lung}$ | liter/hour | 390.000 | 386.343 |
| $Q_{Heart}$ | liter/hour | 15.600 | 15.454 |
| $Q_{Stomach}$ | liter/hour | 3.900 | 3.863 |
| $Q_{Spleen}$ | liter/hour | 11.700 | 11.590 |
| $Q_{Pancreas}$ | liter/hour | 3.900 | 3.863 |
| $Q_{Liver}$ | liter/hour | 99.450 | 98.517 |
| $Q_{Adipose}$ | liter/hour | 19.500 | 19.317 |
| $Q_{Skin}$ | liter/hour | 19.500 | 19.317 |
| $Q_{Muscle}$ | liter/hour | 66.300 | 65.678 |
| $Q_{Bone}$ | liter/hour | 19.500 | 19.317 |
| $Q_{RestOfBody}$ | liter/hour | 29.250 | 28.976 |
| $Q_{Kidney}$ | liter/hour | 74.100 | 73.405 |
| $Q_{PortalVein}$ | liter/hour | 74.100 | 73.405 |
| $Q_{HepaticArtery}$ | liter/hour | 25.350 | 25.112 |
| $MPPGL$ | milligram/gram | 40.161 | 39.791 |
| $W_{Liver}$ | gram | 1726.027 | 1718.567 |
| $Q_{Duodenum}$ | liter/hour | 3.370 | 3.323 |
| $Q_{Jejunum}$ | liter/hour | 16.910 | 16.767 |
| $Q_{Ileum}$ | liter/hour | 18.720 | 18.544 |
| $Q_{LargeIntestine}$ | liter/hour | 15.600 | 15.454 |
| $CYP1A2_{Abundance,Liver}$ | picomole/milligram | NA | 36.400 |
| $CYP2E1_{Abundance,Liver}$ | picomole/milligram | NA | 61.000 |
| $CPPGL$ | milligram/gram | 80.000 | NA |
| **For 33 ODE-system with oral dosing** | | | |
| $Dissolution_Parameter$ | liter/(milligram*hour) | 0.076 | NA |
| $Solubility_{Duodenum}$ | milligram/liter | 13661.889 | |
| $Solubility_{Ileum}$ | milligram/liter | 13709.584 | |
| $Solubility_{Jejunum}$ | milligram/liter | 13668.842 | |
| $Solubility_{LargeIntestine}$ | milligram/liter | 13664.967 | |
| $Solubility_{Stomach}$ | milligram/liter | 13650.000 | |
| $ka_{Duodenum}$ | 1/hour | 3.646 | |
| $ka_{Ileum}$ | 1/hour | 5.760 | |
| $ka_{Jejunum}$ | 1/hour | 4.937 | |
| $ka_{LargeIntestine}$ | 1/hour | 3.570 | |
| $ka_{Stomach}$ | 1/hour | 4.320 | |
| $kt_{Duodenum}$ | 1/hour | 3.125 | |
| $kt_{Ileum}$ | 1/hour | 0.455 | |
| $kt_{Jejunum}$ | 1/hour | 0.676 | |
| $kt_{LargeIntestine}$ | 1/hour | 0.028 | |
| $kt_{Stomach}$ | 1/hour | 0.857 | |